\documentclass[10pt,twocolumn,letterpaper]{article}

\usepackage[pagenumbers]{cvpr} % To force page numbers, e.g. for an arXiv version

\usepackage{amssymb}
\usepackage{amsmath,amsfonts}
\usepackage{amsopn}
\usepackage{bm} %
\usepackage{multirow}
\usepackage{tabularx}
\usepackage{float} 
\newcommand{\vct}[1]{\boldsymbol{#1}} %
\newcommand{\mat}[1]{\boldsymbol{#1}} %

\newcommand{\field}[1]{\mathbb{#1}}

\newcommand{\R}{\field{R}} %

\newcommand{\ProbOpr}[1]{\mathbb{#1}}

\newcommand{\expect}[2]{%
\ifthenelse{\equal{#2}{}}{\ProbOpr{E}_{#1}}
{\ifthenelse{\equal{#1}{}}{\ProbOpr{E}\left[#2\right]}{\ProbOpr{E}_{#1}\left[#2\right]}}} %
\newcommand{\var}[2]{%
\ifthenelse{\equal{#2}{}}{\ProbOpr{VAR}_{#1}}
{\ifthenelse{\equal{#1}{}}{\ProbOpr{VAR}\left[#2\right]}{\ProbOpr{VAR}_{#1}\left[#2\right]}}} %

\newcommand{\vx}{{\vct{x}}}
\newcommand{\vy}{\vct{y}}

\newcommand{\vw}{\vct{w}}

\newcommand{\mA}{\mat{A}}

\newcommand{\mH}{\mat{H}}
\newcommand{\mS}{\mat{S}}

\newcommand{\mL}{\mat{L}}

\newcommand{\eat}[1]{}

\newcommand{\Ours}{{Finer-CAM}\xspace}

\usepackage{multicol}
\usepackage{multirow}
\usepackage{cuted}

\definecolor{cvprblue}{rgb}{0.21,0.49,0.74}
\usepackage[pagebackref,breaklinks,colorlinks,allcolors=cvprblue]{hyperref}
\newcommand{\magnify}[0]{\raisebox{-0.005\linewidth}{\includegraphics[width=.035\textwidth]{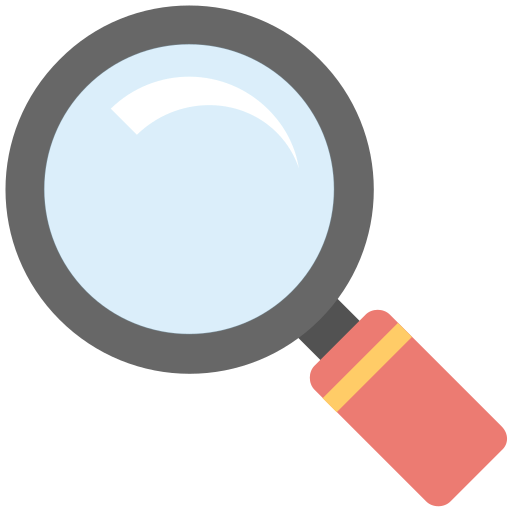}}}

\title{Finer-CAM~\magnify~: Spotting the Difference\\ Reveals Finer Details for Visual Explanation}

\author{Ziheng Zhang\thanks{Equal contributions}\hspace{4pt}, Jianyang Gu$^*$, Arpita Chowdhury, Zheda Mai, David Carlyn,\\ Tanya Berger-Wolf, Yu Su, Wei-Lun Chao
\\
\vspace{-5pt}
\\ 
 The Ohio State University %\\
}
\begin{document}

\maketitle

\begin{strip}
\vskip -35pt
%\begin{figure*}[!hb]
\centering
\begin{minipage}{0.23\textwidth}
\includegraphics[width=\linewidth]{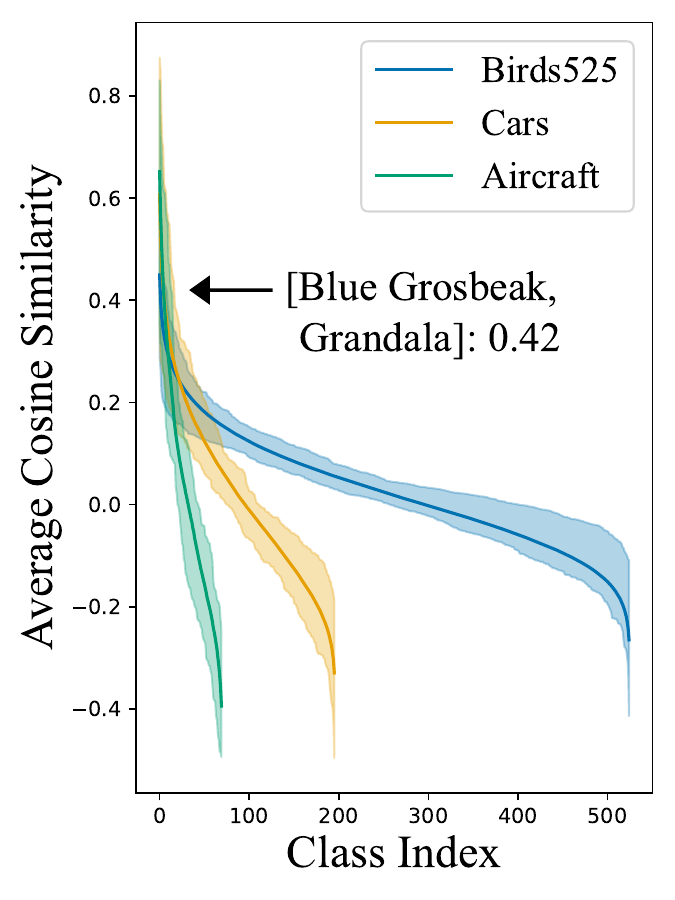}
\end{minipage}
% \hfill
\raisebox{-0.3em}{
\begin{minipage}{0.67\textwidth}
\includegraphics[width=\linewidth]{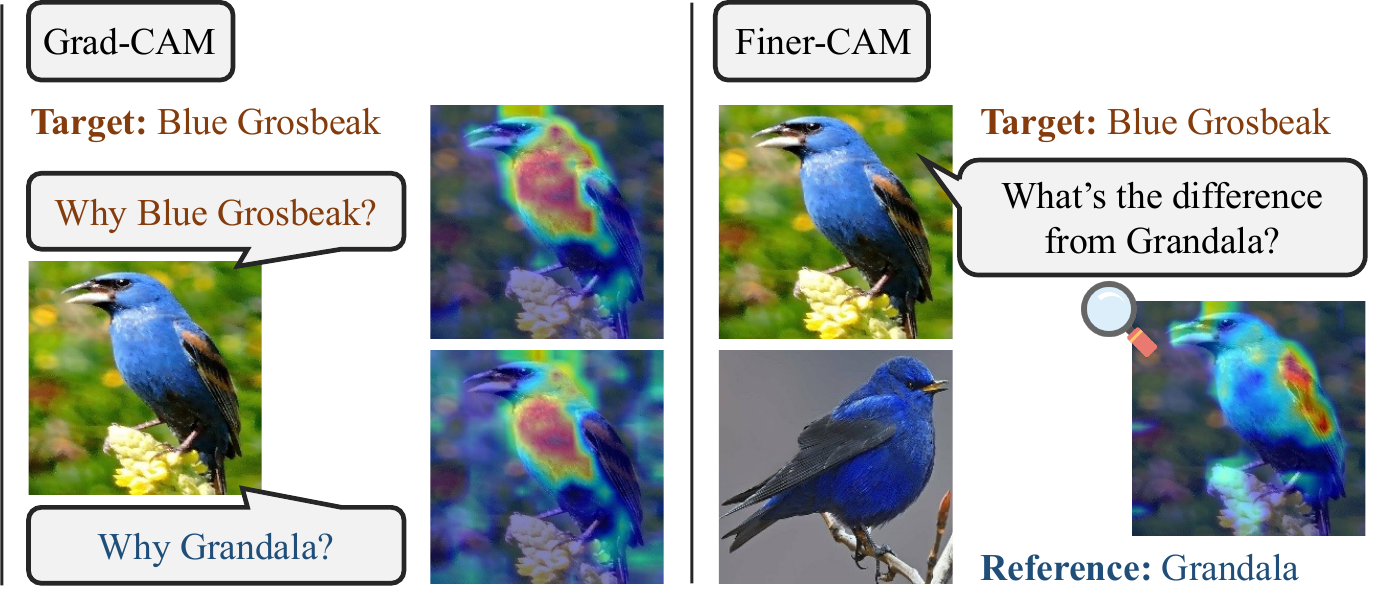}
\end{minipage}
}
\vskip -8pt
\captionof{figure}{\textbf{Illustration of \Ours. }\textbf{Left:} Sorted cosine similarity between linear classifier weights, averaged across all classes (details in the supplementary). Many pairs of classes are highly similar, yet neural networks can effectively distinguish them to achieve high fine-grained classification accuracy.
\textbf{Middle}: Standard CAM methods highlight main regions contributing to the target class's logit value, inadvertently including regions predictive of similar classes and overshadowing fine discriminative details.  \textbf{Right}: We propose \Ours to explicitly compare the target class with similar classes and spot the difference, enabling accurate localization of discriminative details. }
\label{fig:teaser}
\vskip -5pt
%\end{figure*}
\end{strip}

\begin{abstract}
Class activation map (CAM) has been widely used to highlight image regions that contribute to class predictions. Despite its simplicity and computational efficiency, CAM often struggles to identify discriminative regions that distinguish visually similar fine-grained classes. Prior efforts address this limitation by introducing more sophisticated explanation processes, but at the cost of extra complexity. In this paper, we propose \textbf{\Ours}, a method that retains CAM's efficiency while achieving precise localization of discriminative regions. Our key insight is that the deficiency of CAM lies not in ``how'' it explains, but in \textbf{``what'' it explains}. Specifically, previous methods attempt to identify all cues contributing to the target class's logit value, which inadvertently also activates regions predictive of visually similar classes.
By explicitly comparing the target class with similar classes and spotting their differences, \Ours suppresses features shared with other classes and emphasizes the unique, discriminative details of the target class.
\Ours is easy to implement, compatible with various CAM methods, and can be extended to multi-modal models for accurate localization of specific concepts. Additionally, \Ours allows adjustable comparison strength, enabling users to selectively highlight coarse object contours or fine discriminative details.
Quantitatively, we show that masking out the top 5\% of activated pixels by \Ours results in a larger relative confidence drop compared to baselines. The source code and demo are available at \href{https://github.com/Imageomics/Finer-CAM}{https://github.com/Imageomics/Finer-CAM}.

\end{abstract} 
    
\section{Introduction}
\label{sec:intro}

% The background of CAM
Deep neural networks can capture texture and structure information in images and leverage these features to recognize the corresponding classes~\cite{he2016deep,dosovitskiy2021an,deng2009imagenet,ridnik2021imagenet21k}. 
Thanks to large-scale datasets and robust training processes, deep learning algorithms have achieved classification accuracies surpassing those of human experts~\cite{he2015delving,szegedy2015going}.
With these advancements comes a growing interest in understanding the mechanisms behind the successful classification of images.
Gaining insight into how specific features influence predictions not only clarifies the decision-making process but also enhances the model's explainability~\cite{arrieta2020explainable,chefer2021transformer,minh2022explainable,escalante2018explainable}.

%for this purpose
A popular method is class activation map (CAM), which employs a linear combination of feature activation maps to highlight image regions contributing to the prediction~\cite{zhou2016learning,selvaraju2017grad,wang2020score}. CAM is easy to implement, compatible with various neural network architectures, and computationally efficient~\cite{oh2021evet}. 
However, it often struggles to identify discriminative details in fine-grained classification tasks.
For example, in~\cref{fig:teaser} (\textit{right}), the major difference between Blue Grosbeak and Grandala lies in the color of their wings, yet Grad-CAM~\cite{selvaraju2017grad} focuses mainly on the body part. 
This localization deficiency is commonly attributed to CAM's one-pass explanation process, which is unable to polish the details.
Recent works have thus explored more sophisticated iterative perturbation~\cite{fong2017interpretable,ribeiro2016should,oh2021evet} or interpretable models~\cite{chefer2021transformer,paul2024simple,xue2022protopformer} to improve the fine-grained explanation.

In this paper, we argue that CAM's deficiency arises not from its explanation process but from \textbf{``what'' it explains}. 
Ideally, in fine-grained tasks, explanation methods should \textit{emphasize the distinctions between visually similar classes}. 
However, standard CAM methods explain the prediction for each class independently, attempting to highlight all regions associated with the category. This inevitably results in coarser activation maps, sometimes covering entire objects.

More specifically, CAM aims to activate regions with features that correlate with the fully connected classifier weights and contribute to high logit values~\cite{selvaraju2017grad}.
We visualize the sorted similarity between linear classifier weights in~\cref{fig:teaser} (\textit{left}).
While the overall similarity is low after model training, certain class pairs still yield high similarity, indicating that they share some common features. 
For instance, Blue Grosbeak and Grandala share similar body features, differing primarily in subtle details, such as wing color.
When Blue Grosbeak (the true class) is used independently as the explanation target in~\cref{fig:teaser} (\textit{middle}), the high similarity in classifier weights results in activations nearly identical to those for Grandala (a similar class).
We thus propose an alternative reason for CAM's poor fine-grained capability: as it focuses solely on the features predictive of the target class, CAM overlooks the fact that these features may also increase the prediction logits for similar classes.

\begin{figure}[t]
    \centering
\includegraphics[width=0.9\linewidth]{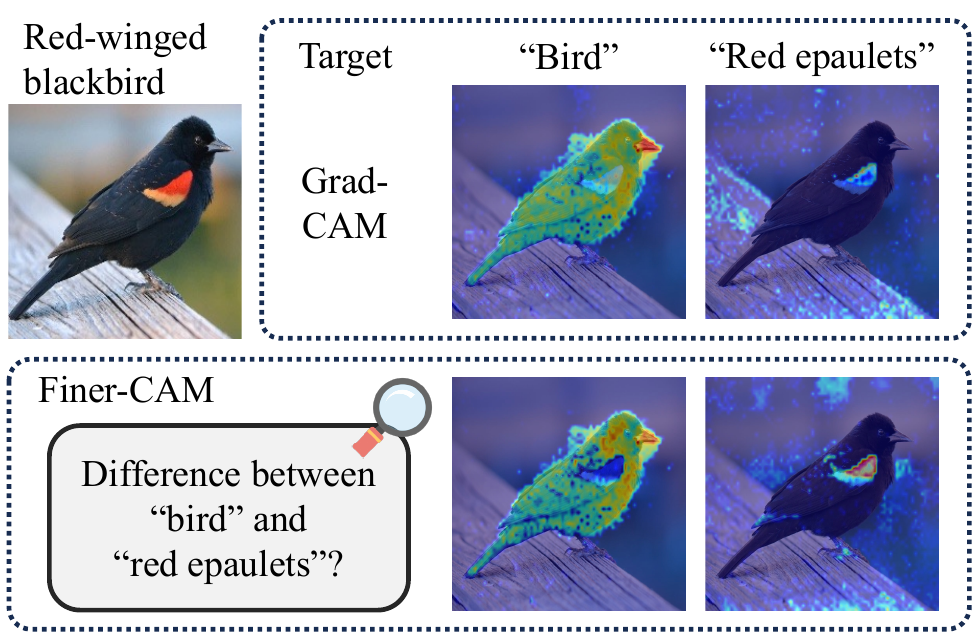}
\vskip -6pt
    \caption{\textbf{\Ours can be extended to multi-modal zero-shot models} to accurately highlight or mask out specific concepts. }
    \vskip -10pt
    \label{fig:teaser-right}
\end{figure}

% Our proposal - highlight the discriminative areas
Intuitively, discriminative regions are easier to identify when similar images are provided for reference, much like a spot-the-difference game.
\emph{Inspired by this, we propose \textbf{\Ours}, a method that explicitly compares the target class with similar classes to reveal the most discriminative feature channels.} 
\Ours is extremely easy to implement and compatible with various CAM-based methods.  
It simply requires changing the explanation target---from the target class's logit value to its difference with a reference class's logit---applicable to both gradient- and score-based CAM methods.
This comparison effectively uncovers key distinctions between similar classes and highlights the corresponding regions in the images. 
Notably, \Ours does not require a reference image to be provided. 
As shown in~\cref{fig:teaser} (\textit{right}), using Grandala's linear classifier weights, not its image, as a reference, \Ours successfully identifies the difference in the wings. 

Beyond compatibility, \Ours's \emph{spot-the-difference} mechanism provides several flexibilities for fine-grained analysis.
First, it supports comparisons with different reference classes, enabling \Ours to identify distinct discriminative regions for the target class, as shown in~\cref{fig:compare-target}. This is particularly valuable in biological domains, where distinguishing different pairs of visually similar species often requires unique traits.
Second, \Ours allows users to adjust the comparison strength---ranging from focusing on the target class's logit alone to emphasizing its difference from a reference class. 
This flexibility enables selective highlighting of coarse object regions or fine discriminative details.
Finally, this adjustability facilitates weighted aggregation of comparisons across multiple similar classes, producing more comprehensive and discriminative activation maps for the target class.
Quantitatively, we show that masking the top 5\% of activated pixels identified by \Ours leads to a greater relative confidence drop in fine-grained classification compared to baselines, demonstrating its effectiveness in localizing discriminative regions.

% Out effects
Additionally, \Ours applies not only to classifiers but also to multi-modal zero-shot models~\cite{radford2021learning}.
As shown in~\cref{fig:teaser-right}, comparing ``red epaulets'' with ``bird'' achieves better localization of the concept on the bird than using only the prompt ``red epaulets.'' 
This extension further enables verification of whether the regions highlighted by class differences align with expert-identified nameable attributes.
 
In summary, \Ours offers a more nuanced understanding of model predictions than previous CAM methods while retaining CAM's computational efficiency.

\section{Related Work}
\label{sec:related}

\subsection{Explainable AI}
Explainable AI (XAI) aims to understand the decision of complex black-box models, and \textit{saliency maps} have been one of the promising venues to provide reasonable explanations. 
Method designs include local optimization~\cite{ribeiro2016should,lundberg_unified_2017}, occlusion-based~\cite{petsiuk2018rise,fong2017interpretable,fong2019understanding}, gradient-based~\cite{baehrens2010explain,rebuffi2020there,smilkov2017smoothgrad}, and CAM-based~\cite{zhou2016learning,selvaraju2017grad,wang2020score,muhammad2020eigen,ramaswamy2020ablation} methods. 
There is also a series of works dedicated to designing evaluation metrics for XAI methods~\cite{yeh2019fidelity,klein2024navigating,chalasani2020concise}.
Among these XAI solutions, we mainly focus on the CAM-based approaches.

\noindent
\textbf{Class activation map (CAM)} uses a linear combination of feature activation maps to illustrate the salient image regions for a target class~\cite{zhou2016learning}. 
Existing methods differ by their weight assignments across the activation maps.
Grad-CAM \cite{selvaraju2017grad} applies classification gradients to indicate the importance of each channel, which is later refined by using positive partial derivatives~\cite{chattopadhay2018grad}, introducing extra axioms~\cite{fu2020axiom}, and fusing multiple layers~\cite{jiang2021layercam}. 
Score-CAM directly uses the influence of each activation map on the final prediction as the corresponding weights~\cite{wang2020score}.
CAM-based methods can localize the target object with all contributing parts. 
However, when it comes to fine-grained classification, the capability to identify discriminative details is often limited. 

\begin{figure*}
    \centering
    \includegraphics[width=\textwidth]{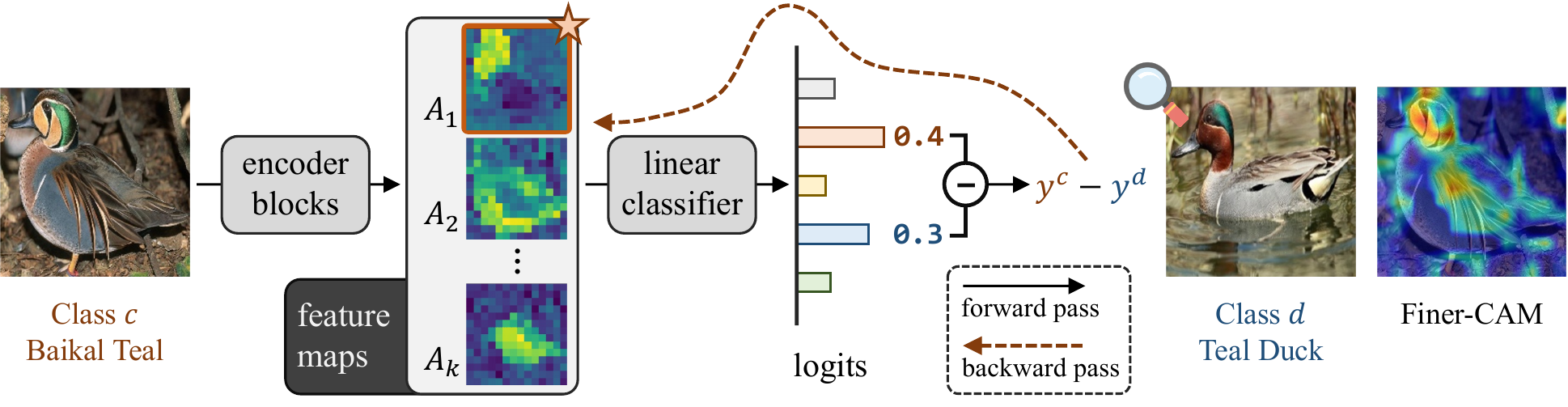}
    \vskip -8pt
    \caption{\textbf{The pipeline of the proposed \Ours method}, with Grad-CAM as the baseline. An image is first passed through the encoder blocks and the subsequent linear classifier to acquire feature maps at the desired network layer and the prediction logits, respectively. Different from standard Grad-CAM, we calculate the gradients of the logit difference between the target class and a visually similar class. In this way, the produced CAM effectively captures and highlights subtle differences between these two classes.}
    \vskip -8pt
    \label{fig:pipeline}
\end{figure*}

\subsection{Fine-grained Classification}
Fine-grained classification aims to distinguish subordinate-level categories within a general category, \eg, different species of birds and models of cars~\cite{wei2021fine,zhao2017survey,wah2011caltech,krause20133d,maji2013fine,piosenka2023birds}. 
Unlike standard image classification, fine-grained tasks often involve subtle differences between classes, localized to specific regions~\cite{zheng2019looking,zheng2017learning}. 
While many cues contribute to class prediction, only a few can deterministically distinguish an object from its visually similar counterparts.
In this work, we focus on spotting the differences between the target class and similar classes and highlight these discriminative details in the images.

\section{Method}
\label{sec:method}

\subsection{Preliminaries}
Class activation map (CAM) is a technique to highlight image regions that contribute to the classification prediction. We denote a neural network classifier by $f: \mathcal{X}\rightarrow \R^C$, which maps the input image $\vx\in\mathcal{X}$ to prediction logits $\vy\in\R^C$.
During the forward pass, a network layer generates $K$ feature maps $\mathcal{A}=\{\mA_i\}_{i=1}^K$ according to the channel number. 
Consider a feature map $\mA_k\in \mathcal{A}$ and the prediction logit $y^c\in \R$ for class $c$. 
CAM assigns an importance weight $\alpha_k^c$ based on the contribution of $\mA_k$ to the prediction logit. The final saliency map $\mL^c$ for class $c$ is produced by a linear combination of the feature maps:

\begin{equation}\label{eq:cam}
    \mL^c=h\left(\sum_k\alpha_k^c \mA_k\right),
\end{equation}
where $h(\cdot)$ is an activation function, typically set as $\texttt{ReLU}$ to focus on features with positive effects on the prediction.  

% of all pixels
Grad-CAM~\cite{selvaraju2017grad} acquires the importance weight based on the average back-propagated gradients with respect to all grids in the feature map:
\begin{equation}\label{eq:gradcam}
    \alpha^c_k=\frac{1}{Z} \sum_i \sum_j \frac{\partial y^c}{\partial A^{ij}_k},
\end{equation}
where $i,j$ represent the feature grid location in $\mA_k$, and $Z$ is the total number of feature grids.
Score-CAM~\cite{wang2020score} obtains $\alpha^c_k$ by measuring the increase of confidence after applying the feature activation map on the original image:
\begin{equation}\label{eq:scorecam}
    \alpha^c_k=f(\vx\circ \mH_k)^c-f(\vx_b)^c,
\end{equation}
where $\vx_b$ is by default a zero input, $\mH_k$ is the upsampled activation map to the original image size, $\circ$ denotes Hadamard Product, and $f(\cdot)^c$ picks the prediction logit for class $c$.

\subsection{Activation via Comparison}
Despite various ways to determine appropriate weights for each feature map, CAM often fails to highlight the most discriminative regions in fine-grained classification tasks.
In such tasks, the distinctions among similar classes are often located in subtle details, whereas CAM tends to activate across the entire object. 
We aim to understand this finding.

Consider the case where Grad-CAM is applied to the last network layer before the linear classifier. 
It has been proved that $\alpha^c_k$ equals the corresponding classifier weight $w^c_k$ that transfers the (averagely pooled) feature map to the prediction logit up to a proportionality constant ($1/Z$)~\cite{selvaraju2017grad}:
\begin{equation} 
w^c_k=\sum_i\sum_j\frac{\partial y^c}{\partial A^{ij}_k}.
\end{equation}
That is, the importance score of the $k$-th channel is exactly the corresponding linear classifier weight. 
As suggested in~\cref{fig:teaser} (\textit{left}), several fine-grained class pairs possess high similarity.
When CAM solely considers the target class $c$ for highlighting regions, it overlooks the fact that the corresponding features may also be predictive of similar classes.

\cref{fig:teaser} (\textit{middle}) illustrates this phenomenon. When CAM is applied to a Blue Grosbeak image, the resulting saliency maps for the true class and Grandala---a similar class---are nearly identical.
The blue body color not only contributes to the correct prediction but also increases the logit for Grandala, as both species share this feature. Consequently, solely explaining the target class inadvertently limits CAM from spotting discriminative regions.

Intuitively, identifying discriminative regions in an image becomes easier when similar references are provided, akin to a spot-the-difference task. 
Inspired by this, we propose \Ours, which assigns activation weights by explicitly comparing the target class with similar ones. 
We first use gradient-based CAM methods to demonstrate the idea. \cref{fig:pipeline} shows the pipeline of \Ours. 

\noindent
\textbf{Gradient-based \Ours.}
The original Grad-CAM only considers the prediction logit of the target class.
We propose to additionally involve similar classes and calculate the gradients of the logit difference:
\begin{equation}\label{eq:finercam}
    \alpha^{c,d}_k=\frac{1}{Z} \sum_i \sum_j \frac{\partial (y^c-\gamma \times y^d)}{\partial A^{ij}_k},
\end{equation}
where $y^d$ is the prediction logit of a similar class $d$, and $\gamma$ is the comparison strength coefficient. Based on the differentiation linearity, we can decompose the partial derivatives:
\begin{equation}
    \frac{\partial (y^c - \gamma \times y^d)}{\partial A^{ij}_k} = \frac{\partial y^c}{\partial A^{ij}_k} - \gamma \times \frac{\partial y^d}{\partial A^{ij}_k}.
\end{equation}
Following the definition in~\cref{eq:gradcam}, we obtain:
\begin{equation}\label{eq:subtraction}
    \alpha^{c,d}_k=\alpha^c_k-\gamma \times \alpha^d_k,
\end{equation}
which we then use to replace $\alpha^c_k$ in~\cref{eq:cam}.

In short, instead of merely capturing features predictive of class $c$ in isolation, the proposed \Ours identifies those that positively contribute to class $c$ while negatively (or less strongly) contributing to class $d$. In~\cref{fig:teaser}, the blue body is a shared trait between both species and does not aid in differentiation. Therefore, it is less activated after the comparison in \Ours.

\noindent
\textbf{Aggregation.}
By controlling the comparison strength $\gamma$, it is possible to adjust the distribution in the saliency map. 
When $\gamma=0$, \Ours degenerates to the baseline Grad-CAM and produces a coarse saliency map very much covering the object.
In contrast, a larger $\gamma$ leads to fine-grained activation of details. 
See~\cref{fig:compare-weight} for illustrations. 
With this flexibility, we can also aggregate \Ours with multiple references to form the final saliency map for the target class:
\begin{equation} \label{eq:multiple-agg}
    L^c= \texttt{ReLU}\left(\frac{1}{T}\sum_t\sum_k\alpha_k^{c,t} \mA_k\right),
\end{equation}
where $T$ is the number of compared reference classes.
The aggregation fuses the key distinctions between the target class and multiple similar classes, making the produced saliency map more comprehensive. 
Note that given the existence of the $\texttt{ReLU}$ activation, the direct subtraction between two saliency maps cannot yield the same result as \Ours, which is further analyzed in~\cref{sec:analysis}. 

\noindent
\textbf{Score-based \Ours.}
The proposed \Ours can be applied to score-based CAM methods as well. 
Building upon~\cref{eq:scorecam}, we add a negative term to de-emphasize features that positively contribute to a similar reference class $d$.
The resulting activation weights thus highlight feature maps that would enlarge the logit difference between the target class and the reference:
\begin{equation}
\alpha^{c,d}_k=f(\vx\circ \mH_k)^c-\gamma \times f(\vx\circ \mH_k)^d-f(\vx_b)^c,
\end{equation}

\begin{figure*}[t]
\includegraphics[width=\textwidth]{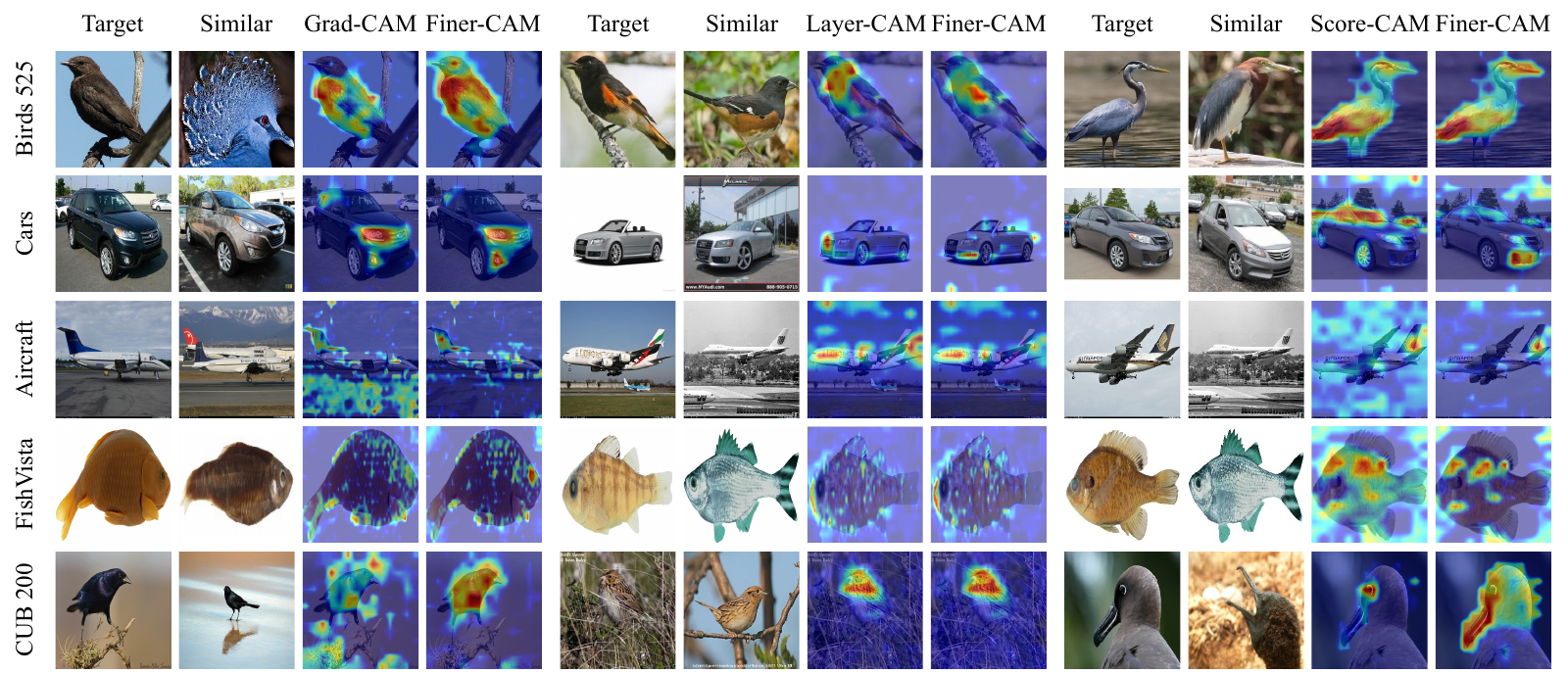}
\vskip-6pt
    \caption{\textbf{The visualization comparison between the proposed \Ours and baseline CAM methods.} For each group, we show the target image, one example image from the most similar class, baseline CAM, and \Ours's results. \Ours localizes and emphasizes the discriminative details, and also suppresses some noise in the baseline CAMs. }\label{fig:samples}
    \vskip -8pt
\end{figure*}

\subsection{Extension to Multi-modal Interaction}
Typically, CAM is applied to explain a classifier on a specific task. 
However, it can also be extended to zero-shot classification scenarios. 
For a pre-trained CLIP model~\cite{radford2021learning}, the fixed linear classifier layer is replaced by text embeddings; the logits are calculated by the similarities between visual and text embeddings. In this setting, CAM highlights image regions that correspond to the semantics of the text prompt.
Similar to the classifier-based scenario, we perform comparisons between different text prompts, enabling more flexible interaction and accurate localization of concepts within the image. 

A valuable application of this extension is verifying the correctness of model activations. 
When both class and attribute labels are provided, \Ours can first be applied to the standard classifier setting to obtain the saliency map that highlights the class difference. 
Then, it can also be used to generate saliency maps for notable attributes (\eg, ``red epaulets'' in~\cref{fig:teaser-right}). 
If the activations for classification align with those for attributes, we consider that the classifier correctly recognizes key traits. 
Conversely, if huge discrepancies arise, the classifier might be biased by other factors, or the provided attribute labels might not be comprehensive enough to distinguish classes. 
Examples are in~\cref{fig:verify}. 

\subsection{Relative Confidence Drop}
There have been a variety of metrics designed to evaluate the faithfulness of XAI methods~\cite{bach2015pixel,petsiuk2018rise,hama2023deletion,klein2024navigating}. 
However, most of them solely focus on the prediction confidence of the target class but overlook its relationship with similar classes. 
As illustrated in~\cref{fig:teaser}, the features contributing to the target class's logit are also predictive of similar classes. 
Therefore, we argue that the masked-out activated regions should degrade the confidence of the target class but have minimal influence on similar classes.
Accordingly, we propose to use the relative confidence drop as the metric in this work.
Given an input image, we first record the initial confidences $p^c$ for the target class and $p^d$ for the most similar class. 
After masking a pre-defined percentage of the most activated pixels, we again acquire the confidences $p_\star^{c}$ and $p_\star^{d}$.
The relative drop is calculated as:
\begin{equation}
    \text{RD}=(p^c-p_\star^{c})-(p^d-p_\star^{d}).
\end{equation}
Larger drops mean that masking the top pixels effectively reduces the confidence in predicting the target class over the reference, indicating more discriminative saliency maps. 

\section{Experiments}
\label{sec:experiments}

\subsection{Implementation Details}
We consider two application scenarios for \Ours, \ie, the standard classifier and multi-modal zero-shot classification settings.
For the classifier setting, we employ a pre-trained CLIP visual backbone and train a linear classifier head on top using the Adam optimizer~\cite{kingma2014adam}. 
The classifier is trained for 100 epochs on each dataset with a learning rate of 3e-4. 
For the zero-shot setting, we directly use the pre-trained CLIP model for inference~\cite{radford2021learning}.
We conduct aggregation over the saliency maps generated by comparing the target class with the top 3 similar classes. 
The weight $\gamma$ in~\cref{eq:finercam} is defaulted as $0.6$, unless stated otherwise.

\begin{table*}[ht!]
    \caption{The quantitative evaluation results on the proposed \Ours and baseline CAM methods. The abbreviations Del., RD., and Loc.~stand for deletion, relative drop, and localization, respectively. The best result in each group and column is highlighted in bold.}
    \label{tab:metric}
    \centering
    \small
    \vskip -4pt
    \resizebox{\linewidth}{!}{
    \begin{tabular}{l|ccc|cccc|cccc}
    \toprule
        \multirow{2}{*}{Method} & \multicolumn{3}{c|}{Birds525} & \multicolumn{4}{c|}{CUB} & \multicolumn{4}{c}{Cars} \\
        & Del. $\downarrow$& RD.$_{\text{@}0.05}$ $\uparrow$ & RD.$_{\text{@}0.1}$ $\uparrow$ & Del. $\downarrow$ & RD.$_{\text{@}0.05}$ $\uparrow$ & RD.$_{\text{@}0.1}$ $\uparrow$ & Loc. $\uparrow$ & Del. $\downarrow$ & RD.$_{\text{@}0.05}$ $\uparrow$ & RD.$_{\text{@}0.1}$ $\uparrow$ & Loc. $\uparrow$ \\
    \midrule
        Grad-CAM~\cite{selvaraju2017grad} &
        0.079 & 0.174 & 0.245 &
        \textbf{0.024} & 0.101 & 0.113 & 0.582 &
        \textbf{0.024} & 0.055 & 0.067 & 0.561 \\
        + Finer &
        \textbf{0.076} & \textbf{0.192} & \textbf{0.260} &
        \textbf{0.024} & \textbf{0.112} & \textbf{0.121} &\textbf{0.629} &
        \textbf{0.024} & \textbf{0.060} & \textbf{0.071} & \textbf{0.565} \\
    \midrule
        Layer-CAM~\cite{jiang2021layercam} &
        \textbf{0.071} & 0.186 & 0.255 &
        \textbf{0.023} & 0.106 & 0.116 & 0.625 &
        \textbf{0.023} & 0.059 & 0.069 & 0.581 \\
        + Finer &
        \textbf{0.071} & \textbf{0.201} & \textbf{0.270} &
        0.024 & \textbf{0.110} & \textbf{0.120} & \textbf{0.682} &
        \textbf{0.023} & \textbf{0.064} & \textbf{0.074} & \textbf{0.592} \\
    \midrule
        Score-CAM~\cite{wang2020score} &
        \textbf{0.088} & 0.151 & 0.217 &
        \textbf{0.029} & 0.090 & 0.102 & 0.670 &
        \textbf{0.027} & 0.051 & 0.061 & 0.565 \\
        + Finer &
        0.089 & \textbf{0.163} & \textbf{0.227} &
        \textbf{0.029} & \textbf{0.098} & \textbf{0.109} & \textbf{0.683} &
        \textbf{0.027} & \textbf{0.054} & \textbf{0.066} & \textbf{0.575} \\
    \bottomrule
    \end{tabular}
    }
    \vskip -8pt
\end{table*}

\noindent
\textbf{Datasets.}
In this paper, we mainly adopt five fine-grained classification datasets covering different general categories including Birds-525~\cite{piosenka2023birds}, CUB-200~\cite{wah2011caltech}, Cars~\cite{krause20133d}, Aircraft~\cite{maji2013fine}, and FishVista~\cite{mehrab2024fish}.
Please refer to the supplementary material for more data details. 

\subsection{Experimental Results}
We first compare the proposed \Ours with different baseline CAM methods including Grad-CAM~\cite{selvaraju2017grad}, LayerCAM~\cite{jiang2021layercam}, and Score-CAM~\cite{wang2020score}. 

\noindent
\textbf{Visualization comparison.}
The visualization of example saliency maps generated by baseline methods and our proposed \Ours is shown in~\cref{fig:samples}.
\Ours shows advantages over baseline CAM methods in three aspects.
First, when baseline CAM methods focus on regions that also contribute to predicting similar classes, \Ours localizes the discriminative details by explicitly spotting the difference from those classes. 
Second, \Ours emphasizes the key regions with noticeably higher activations. 
Especially for the Fish-Vista dataset, the key distinctions (eye and lip in the first two examples) are assigned with high importance by \Ours. 
Last, \Ours suppresses activation noises in the backgrounds.
In the Grad-CAM-CUB example, the background beach environment might bring biases for predicting the target class. 
When the compared classes correlate with a similar background, \Ours effectively suppresses the activation on the background but focuses sharply on the object. 

\begin{figure}[t]
    \centering
\begin{subfigure}[t]{0.46\columnwidth}
    \includegraphics[width=\textwidth]{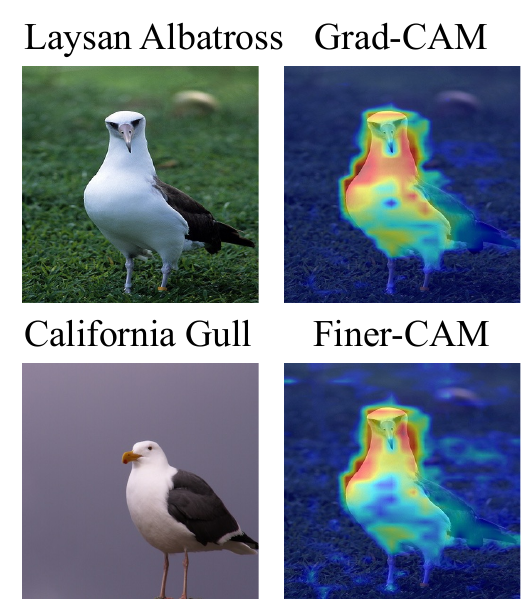}
\end{subfigure}
\begin{subfigure}[t]{0.53\columnwidth}
\raisebox{-0.5em}{
    \includegraphics[width=\textwidth]{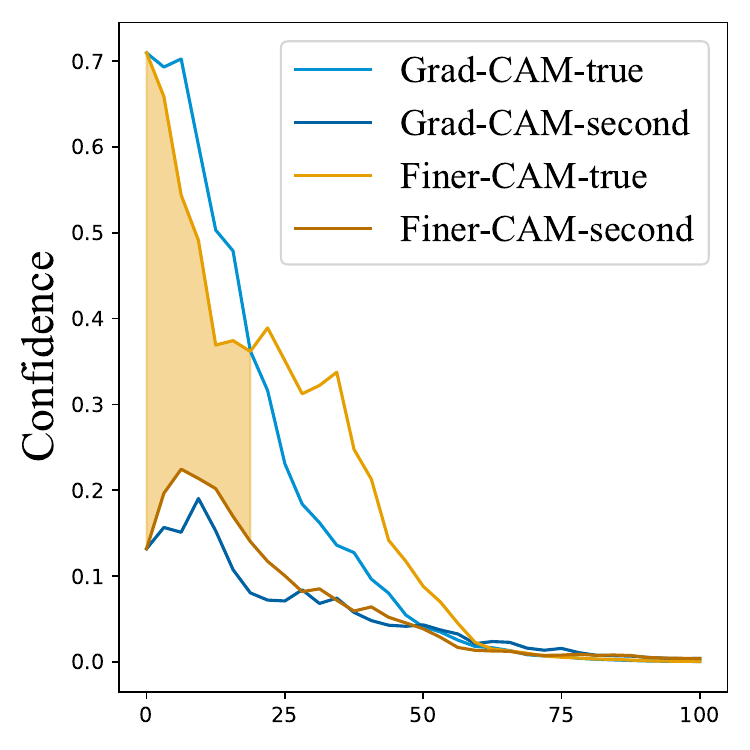}
}
\end{subfigure}
\begin{subfigure}[t]{0.46\columnwidth}
    \includegraphics[width=\textwidth]{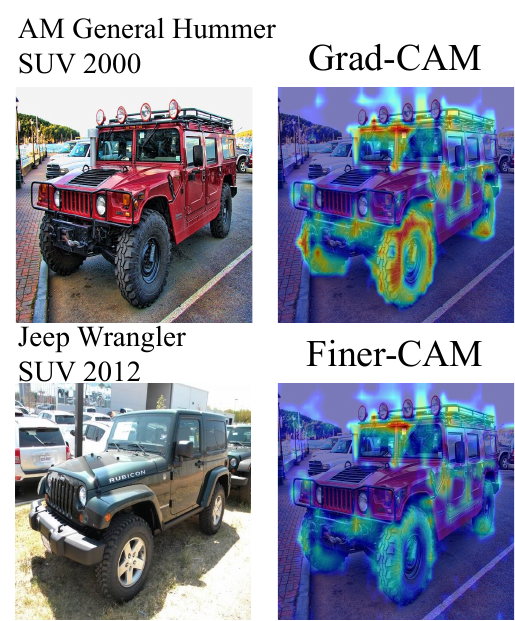}
\end{subfigure}
\begin{subfigure}[t]{0.53\columnwidth}
\raisebox{-0.6em}{
    \includegraphics[width=\textwidth]{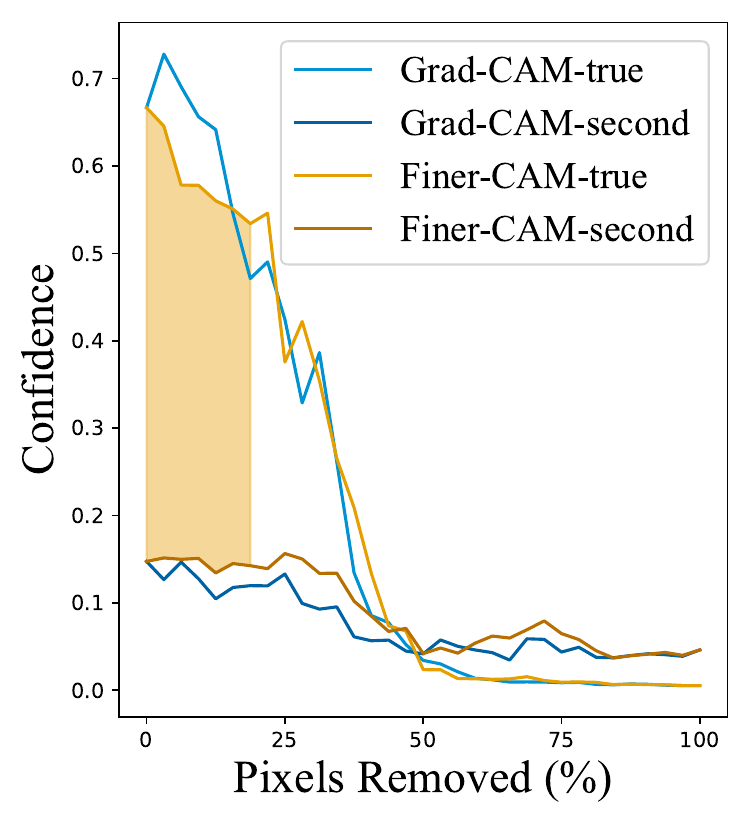}
}
\end{subfigure}
\vskip -5pt
    \caption{\textbf{The saliency maps by Grad-CAM and \Ours with deletion curves.} In each group, the top-left is the target image, while the bottom-left is an example image from the most similar class. In addition to the prediction confidence of the target class, we also show the curve of the second predicted class.}
    \label{fig:deletion}
    \vskip -12pt
\end{figure}

\noindent
\textbf{Deletion curve.}
When the proposed \Ours spots discriminative regions in the images, a feasible way to evaluate the efficacy is the deletion curve~\cite{petsiuk2018rise}. 
In~\cref{fig:deletion}, we show the deletion curve for two example images. 
Different from the standard usage where only the prediction confidence of the target class (denoted as ``true'' in the figure) is considered, we additionally show the deletion curve of the second predicted class. 
Ideally, the activated discriminative regions should only represent the target class. Therefore, removing these pixels should diminish the prediction gap between the target class and similar classes.
We demonstrate that the proposed \Ours yields a smaller gap (the colored area in the figure) when masking out the top 20 percent of activated pixels, compared with the Grad-CAM baseline. 

Dataset-wise, in~\cref{tab:metric}, we present the average deletion AUC results on Birds-525~\cite{piosenka2023birds}.
However, there only exist negligible differences between baselines and \Ours. 
We argue that firstly, as the activated regions are often fine-grained details, it is meaningless to compare the whole deletion curve.
In addition, the deletion curve only focuses on the target class prediction, but overlooks the relationship between similar classes. 
Therefore, we further apply the relative drop metric to illustrate the advantage of the proposed \Ours method. 

\begin{figure*}[t]
    \centering
    \begin{subfigure}[c]{0.63\textwidth}
    \includegraphics[width=\linewidth]{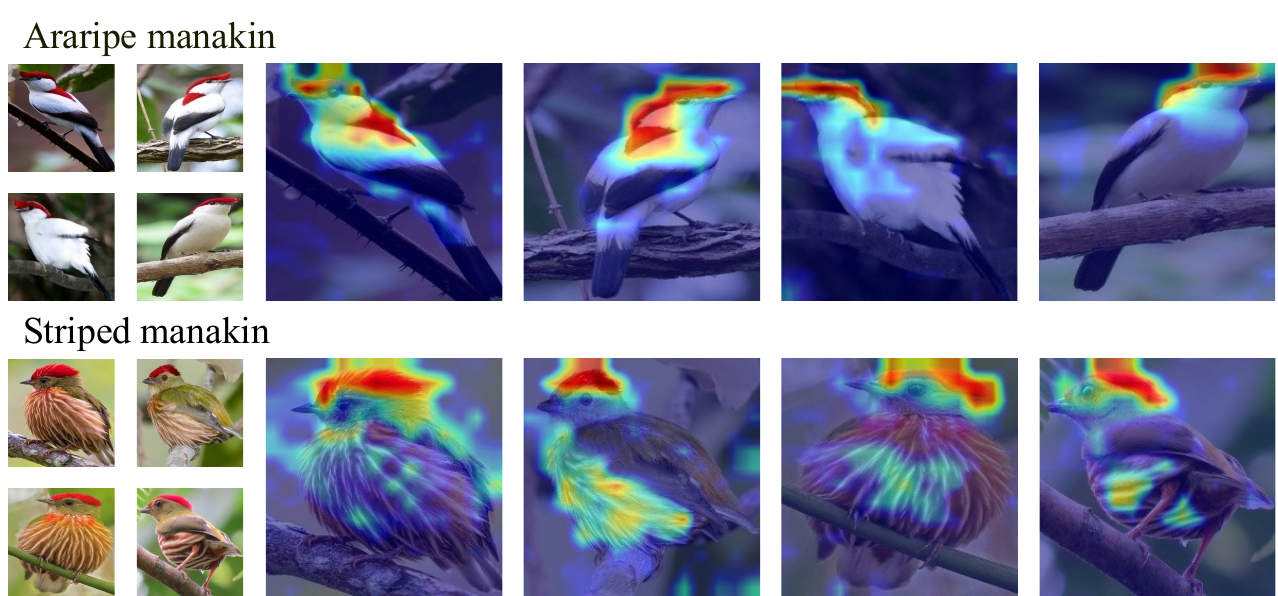}
    \caption{}
    \label{fig:consistency}
    \end{subfigure}
    \raisebox{-0.7em}{
    \begin{subfigure}[c]{0.34\textwidth}
    \includegraphics[width=\linewidth]{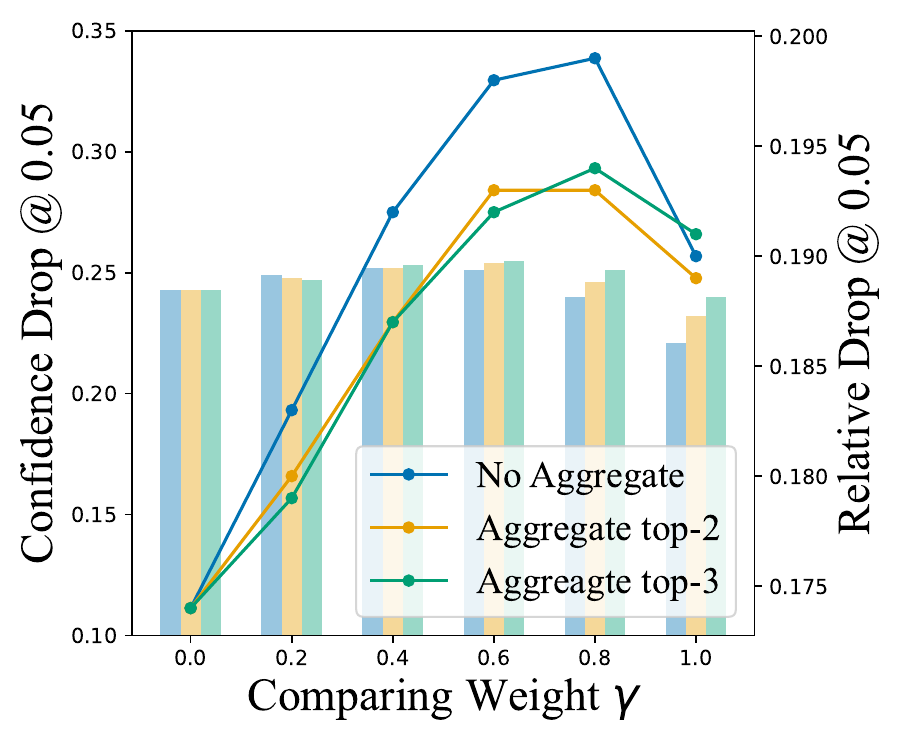}
    \vspace{-1.8em}
    \caption{}
    \label{fig:weight}
    \end{subfigure}
    }
    \vskip -8pt
    \caption{(a) Among images of the same species, \Ours can generate saliency maps that consistently emphasize the same traits. (b) The confidence drop (bars) and relative confidence drop (curves) results when masking out the top 5\% activated pixels with different comparison strength $\gamma$ and aggregation strategies. }
    \vskip -10pt
\end{figure*}

\noindent
\textbf{Relative drop.}
Given that the activated regions are densely located at discriminative regions, we only consider comparing the relative drop when removing the first 5\% and 10\% of total pixels, which is denoted as RD.$_{\text{@}0.05}$ and RD.$_{\text{@}0.1}$, respectively. The results are shown in~\cref{tab:metric}. 
When standard deletion AUC yields similar results, the proposed \Ours provides a stable improvement over baselines on the relative drop metric. 
The results suggest that the regions highlighted by \Ours are more discriminative for recognizing the target class. 

\noindent
\textbf{Localization.}
Through comparison, \Ours illustrates better localization capability for fine-grained details. 
Accordingly, we run the energy-based pointing game~\cite{wang2020score} to provide quantitative evaluation, where activations are expected to be distributed inside the bounding boxes of target objects. 
Based on the annotation availability, the experiments are conducted on CUB-200~\cite{wah2011caltech} and Cars~\cite{krause20133d}. The results are shown in~\cref{tab:metric}, which suggest a substantial advantage of the proposed \Ours over the compared baselines.
The higher pointing game scores also indicate better suppression effects of \Ours on the noises. 

\noindent
\textbf{Consistency.}
Explainable methods are expected to give consistent explanations for instances within one class~\cite{lundberg2017unified,klein2024navigating}. 
In the context of CAM, different instances are supposed to activate similar features. 
We visualize some example saliency maps for images within the same class in~\cref{fig:consistency}.
Although the images present different poses, \Ours consistently highlights the discriminative image regions. 
We also include the comparison with more XAI methods and additional results on more network backbones in the supplementary material. 

\subsection{Analysis and Discussion}
\label{sec:analysis}
In this section, we conduct further analysis of the proposed \Ours method. The method is by default applied on Grad-CAM, which is identical to the ``Fine-Grad-CAM'' in the last section.
Birds-525 dataset is adopted for analysis.

\begin{figure}[t]
    \centering
    \includegraphics[width=\columnwidth]{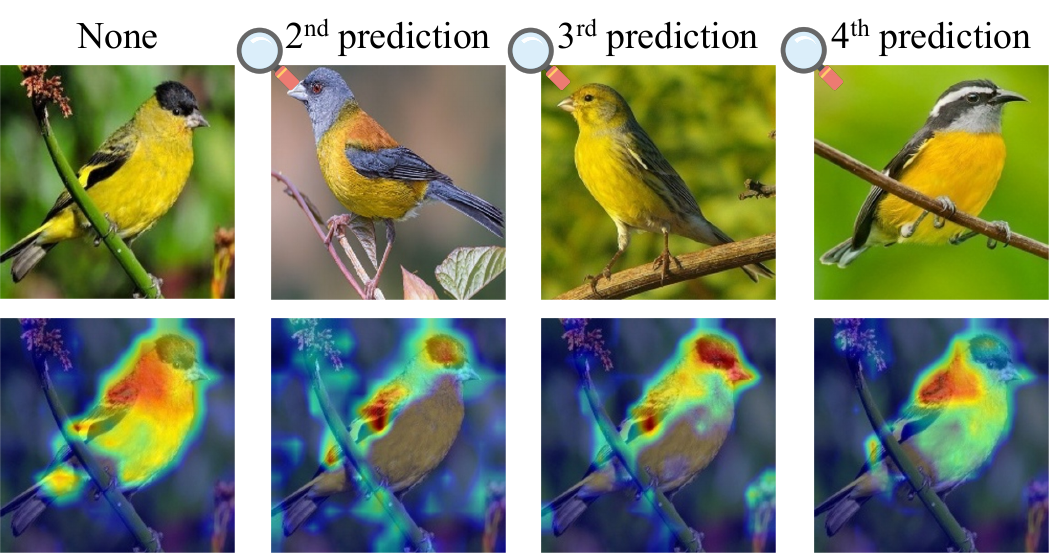}
    \vskip-5pt
    \caption{\textbf{Comparing the target class with different similar classes leads to a variety of activations.} ``None'' denotes the baseline Grad-CAM without comparison.  The class names are: Andean Siskin, Patagonian Sierra Finch, Canary, and Bananaquit.}
    \label{fig:compare-target}
    \vskip -6pt
\end{figure}

\begin{table}[t]
    \caption{Quantitative results of comparing the target class with different references. Del. and RD.$_{\text{@}0.05}$ represent deletion AUC and relative drop when masking out the top 5\% activated pixels.}
    \vskip-5pt
    \label{tab:target}
    \centering
    \small
    \resizebox{\columnwidth}{!}{
    \begin{tabular}{l|ccccc}
    \toprule
        \multirow{2}{*}{Metric} & \multicolumn{5}{c}{Comparing Target} \\
        & None & 2nd Pred & 3rd Pred & 4th Pred & Aggre. \\
    \midrule
        Del. $\downarrow$ & 0.079 & 0.079 & 0.080 & 0.081 & \textbf{0.076} \\
        RD.$_{\text{@}0.05}$ $\uparrow$ & 0.174 & \textbf{0.198} & 0.178 & 0.174 & 0.192 \\
    \bottomrule
    \end{tabular}
    }
    \vskip -6pt
\end{table}

\noindent
\textbf{Comparison reference.}
By comparing the target class with a similar class, \Ours spots the discriminative regions in the images, where the activation map is dependent on the difference between these two classes. 
Therefore, when changing the comparison reference, \Ours will focus on different cues for the target class. 
We first visualize the produced activation maps in~\cref{fig:compare-target}.
The original Grad-CAM highlights the yellow neck and chest regions, but the similar classes also possess yellow body parts. 
Thus, by conducting comparisons, \Ours spots the wing, head, and back for these top 3 similar classes, respectively. 

We also conduct quantitative analysis to investigate the influence of the comparison references using the deletion and relative drop metrics in~\cref{tab:target}. 
Comparing the target class with the second predicted class produces direct optimization of the relative drop metric, while the aggregation of all the similar classes yields the best results in general.

\begin{figure}[t]
    \centering
    \includegraphics[width=\columnwidth]{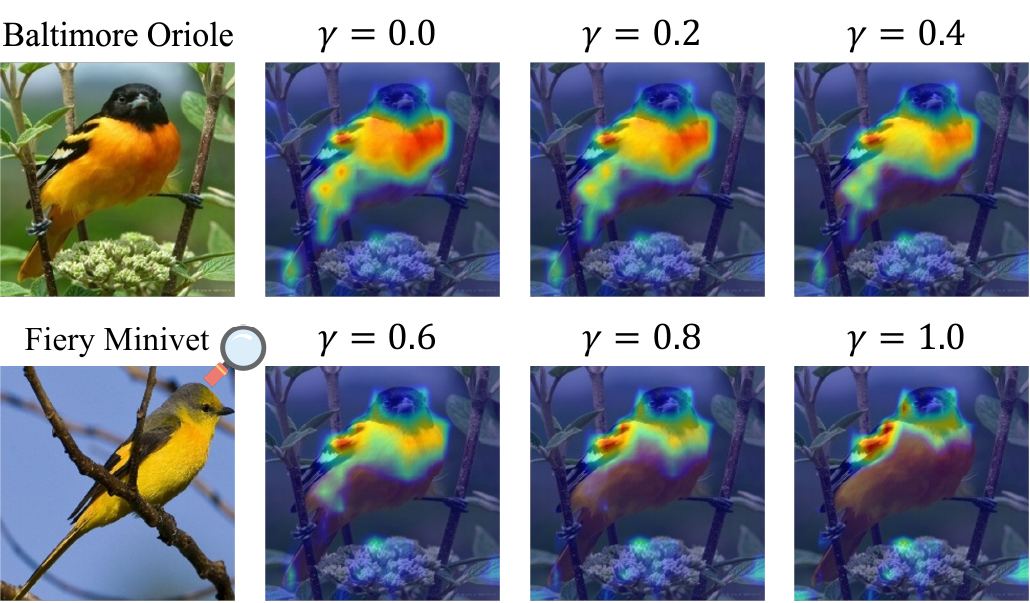}
    \vskip-5pt
    \caption{With a large comparison strength $\gamma$, the activation only focuses on fine-grained details. On the contrary, a small $\gamma$ leads to coarse activation covering the entire object. }
    \label{fig:compare-weight}
    \vskip -8pt
\end{figure}

\begin{figure*}[t]
    \begin{subfigure}[t]{0.7\textwidth}
        \includegraphics[width=\textwidth]{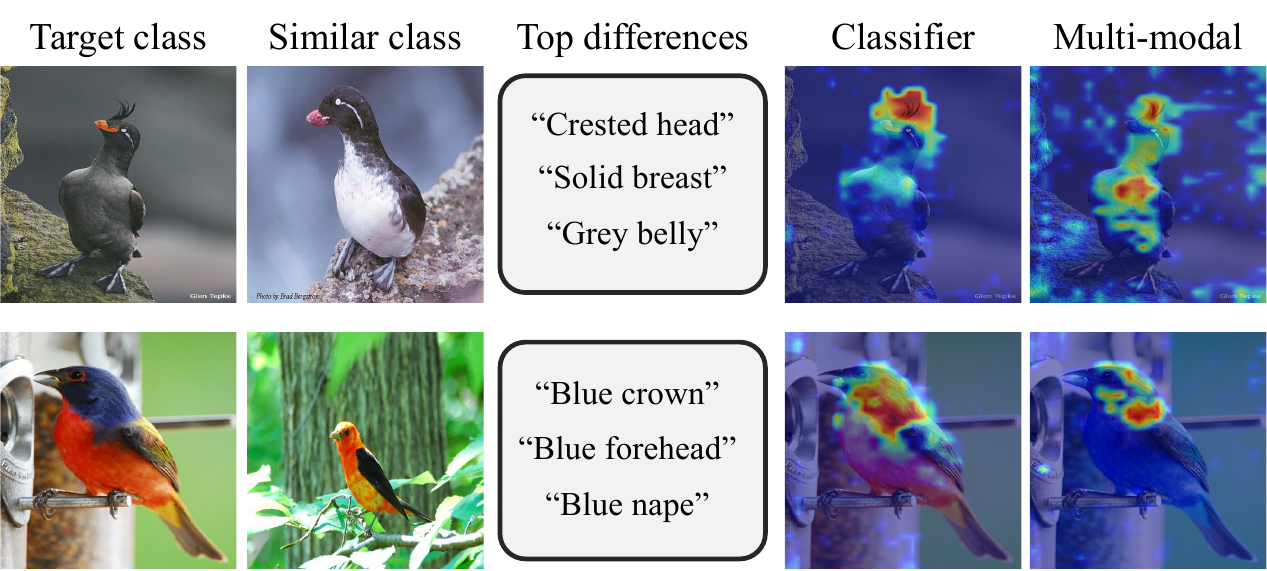}
        \caption{}\label{fig:verify}
    \end{subfigure}
    \hfill
    \begin{subfigure}[t]{0.285\textwidth}
        \includegraphics[width=\textwidth]{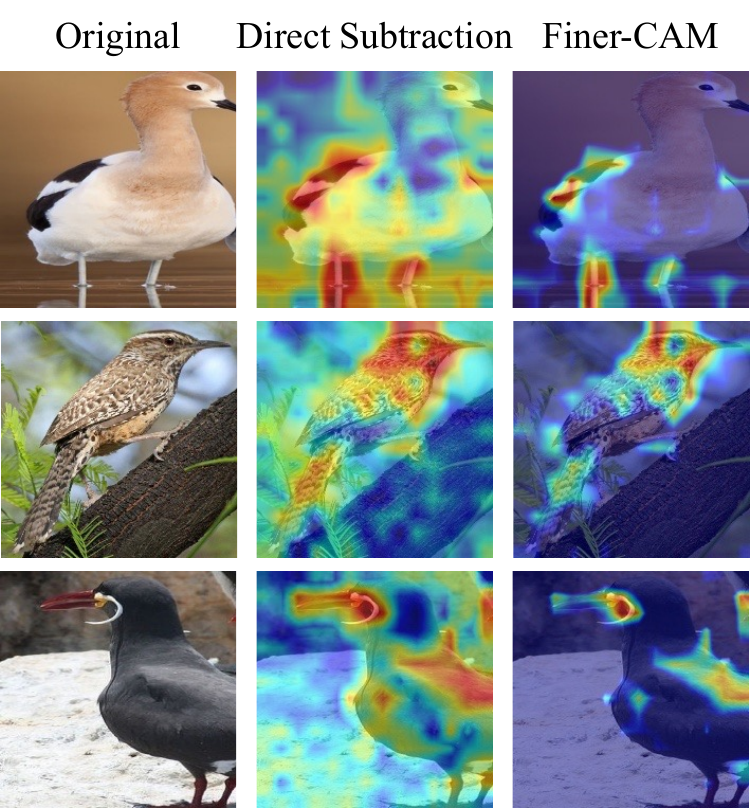}
        \caption{}\label{fig:direct-subtract}
    \end{subfigure}
    \vskip -4pt
    \caption{(a) \Ours can be applied to verify if the classifier learns faithful knowledge of the trait differences between classes. For each row, we show images of the target class and the most similar class, together with the \Ours results generated by the classifier and the multi-modal model. (b) Directly subtracting saliency maps of similar classes from that of the target class will generate noisy results. In contrast, \Ours produces clean backgrounds and highlights fine-grained details.}
    \vskip -4pt
\end{figure*}

\noindent
\textbf{Comparison strength.}
The comparison strength $\gamma$ represents the extent to which we want to suppress the features that also contribute to predicting similar classes. 
As shown in~\cref{fig:compare-weight}, a large $\gamma$ produces saliency maps focusing on fine-grained details, while a small $\gamma$ leads to coarser results that are similar to baseline CAM methods. 

Additionally, we evaluate the influence of the comparison strength by quantitative metrics in~\cref{fig:weight}, where the aggregation of multiple similar classes is also investigated.
As the comparison strength increases, the relative drop is boosted at a fast pace and reaches the peak when $\gamma=0.8$. Further strengthening the comparison leads to slight performance drops.
The absolute confidence drop of the target class shows a similar trend. 
Based on both metrics, we adopt the strength $\gamma=0.6$ and the aggregation of the top 3 similar classes as our final design. 

\noindent
\textbf{Extension to multi-modal scenario.}
We present examples of applying \Ours to the multi-modal scenario in~\cref{fig:teaser-right}.
When Grad-CAM is directly asked to highlight the ``red epaulets,'' it only yields weak activations. 
Comparatively, by comparing the prompts of ``red epaulets'' and ``bird,'' \Ours accurately localizes or masks out the target region. 
This extension provides flexible interaction to activate specific text concepts in the images. 

\noindent
\textbf{Activation faithfulness examination.}
Based on the above extension, \Ours can further be utilized to examine the faithfulness of classifier prediction. 
We experiment with the CUB-200 dataset~\cite{wah2011caltech}, where the attribute differences between classes are provided in annotations.~\cref{fig:verify} presents two examples. In the first row, among the three different attributes, the classifier mainly looks at the crest. The second example emphasizes the blue color around the bird's head, which is also captured by the classifier. The results suggest that the classifier offers faithful prediction consistent with the actual trait differences. More details of the examination are provided in the supplementary. 

\noindent
\textbf{Comparison to CAM subtraction.}
Based on the analysis in~\cref{eq:subtraction}, the proposed \Ours is equivalent to the subtraction of importance weights. 
However, due to the existence of \texttt{ReLU} activation, the result cannot be directly acquired by subtracting the saliency maps of similar classes from that of the target class. 
We illustrate the difference in~\cref{fig:direct-subtract}.
While direct subtraction leads to noisy saliency maps, \Ours produces clean backgrounds and focuses on discriminative regions. 

\section{Conclusion}

We investigate CAM's poor localization capability in fine-grained tasks. We argue that the explanation should not only focus on the target class but also consider visually similar classes to highlight discriminative regions. Accordingly, we propose \Ours, a saliency map approach dedicated to highlighting the regions that differentiate the target class from similar ones. 
\Ours produces accurate activations on fine-grained details, achieving much higher relative confidence drops compared with baseline methods. It can be further extended to multi-modal zero-shot models to accurately activate specific concepts. 
Without sacrificing the simplicity and efficiency of CAM methods, \Ours opens up new possibilities for explainable AI venues.

\section*{Acknowledgment}
This research is supported in part by grants from the National Science Foundation (OAC-2118240, HDR Institute: Imageomics). The authors are grateful for the generous support of the computational resources from the Ohio Supercomputer Center.

% \clearpage
{
    \small
    \bibliographystyle{ieeenat_fullname}
    \bibliography{main}

\begin{thebibliography}{48}
\providecommand{\natexlab}[1]{#1}
\providecommand{\url}[1]{\texttt{#1}}
\expandafter\ifx\csname urlstyle\endcsname\relax
  \providecommand{\doi}[1]{doi: #1}\else
  \providecommand{\doi}{doi: \begingroup \urlstyle{rm}\Url}\fi

\bibitem[Arrieta et~al.(2020)Arrieta, D{\'\i}az-Rodr{\'\i}guez, Del~Ser, Bennetot, Tabik, Barbado, Garc{\'\i}a, Gil-L{\'o}pez, Molina, Benjamins, et~al.]{arrieta2020explainable}
Alejandro~Barredo Arrieta, Natalia D{\'\i}az-Rodr{\'\i}guez, Javier Del~Ser, Adrien Bennetot, Siham Tabik, Alberto Barbado, Salvador Garc{\'\i}a, Sergio Gil-L{\'o}pez, Daniel Molina, Richard Benjamins, et~al.
\newblock Explainable artificial intelligence (xai): Concepts, taxonomies, opportunities and challenges toward responsible ai.
\newblock \emph{Information fusion}, 58:\penalty0 82--115, 2020.

\bibitem[Bach et~al.(2015)Bach, Binder, Montavon, Klauschen, M{\"u}ller, and Samek]{bach2015pixel}
Sebastian Bach, Alexander Binder, Gr{\'e}goire Montavon, Frederick Klauschen, Klaus-Robert M{\"u}ller, and Wojciech Samek.
\newblock On pixel-wise explanations for non-linear classifier decisions by layer-wise relevance propagation.
\newblock \emph{PloS one}, 10\penalty0 (7):\penalty0 e0130140, 2015.

\bibitem[Baehrens et~al.(2010)Baehrens, Schroeter, Harmeling, Kawanabe, Hansen, and M{\"u}ller]{baehrens2010explain}
David Baehrens, Timon Schroeter, Stefan Harmeling, Motoaki Kawanabe, Katja Hansen, and Klaus-Robert M{\"u}ller.
\newblock How to explain individual classification decisions.
\newblock \emph{JMLR}, 11:\penalty0 1803--1831, 2010.

\bibitem[Bousselham et~al.(2024)Bousselham, Petersen, Ferrari, and Kuehne]{bousselham2024grounding}
Walid Bousselham, Felix Petersen, Vittorio Ferrari, and Hilde Kuehne.
\newblock Grounding everything: Emerging localization properties in vision-language transformers.
\newblock In \emph{Proceedings of the IEEE/CVF Conference on Computer Vision and Pattern Recognition}, pages 3828--3837, 2024.

\bibitem[Chalasani et~al.(2020)Chalasani, Chen, Chowdhury, Wu, and Jha]{chalasani2020concise}
Prasad Chalasani, Jiefeng Chen, Amrita~Roy Chowdhury, Xi Wu, and Somesh Jha.
\newblock Concise explanations of neural networks using adversarial training.
\newblock In \emph{ICML}, pages 1383--1391. PMLR, 2020.

\bibitem[Chattopadhay et~al.(2018)Chattopadhay, Sarkar, Howlader, and Balasubramanian]{chattopadhay2018grad}
Aditya Chattopadhay, Anirban Sarkar, Prantik Howlader, and Vineeth~N Balasubramanian.
\newblock Grad-cam++: Generalized gradient-based visual explanations for deep convolutional networks.
\newblock In \emph{WACV}, pages 839--847. IEEE, 2018.

\bibitem[Chefer et~al.(2021)Chefer, Gur, and Wolf]{chefer2021transformer}
Hila Chefer, Shir Gur, and Lior Wolf.
\newblock Transformer interpretability beyond attention visualization.
\newblock In \emph{CVPR}, pages 782--791, 2021.

\bibitem[Deng et~al.(2009)Deng, Dong, Socher, Li, Li, and Fei-Fei]{deng2009imagenet}
Jia Deng, Wei Dong, Richard Socher, Li-Jia Li, Kai Li, and Li Fei-Fei.
\newblock Imagenet: A large-scale hierarchical image database.
\newblock In \emph{CVPR}, pages 248--255. Ieee, 2009.

\bibitem[Dosovitskiy et~al.(2021)Dosovitskiy, Beyer, Kolesnikov, Weissenborn, Zhai, Unterthiner, Dehghani, Minderer, Heigold, Gelly, Uszkoreit, and Houlsby]{dosovitskiy2021an}
Alexey Dosovitskiy, Lucas Beyer, Alexander Kolesnikov, Dirk Weissenborn, Xiaohua Zhai, Thomas Unterthiner, Mostafa Dehghani, Matthias Minderer, Georg Heigold, Sylvain Gelly, Jakob Uszkoreit, and Neil Houlsby.
\newblock An image is worth 16x16 words: Transformers for image recognition at scale.
\newblock In \emph{ICLR}, 2021.

\bibitem[Escalante et~al.(2018)Escalante, Escalera, Guyon, Bar{\'o}, G{\"u}{\c{c}}l{\"u}t{\"u}rk, G{\"u}{\c{c}}l{\"u}, van Gerven, and van Lier]{escalante2018explainable}
Hugo~Jair Escalante, Sergio Escalera, Isabelle Guyon, Xavier Bar{\'o}, Ya{\u{g}}mur G{\"u}{\c{c}}l{\"u}t{\"u}rk, Umut G{\"u}{\c{c}}l{\"u}, Marcel van Gerven, and Rob van Lier.
\newblock \emph{Explainable and interpretable models in computer vision and machine learning}.
\newblock Springer, 2018.

\bibitem[Fong et~al.(2019)Fong, Patrick, and Vedaldi]{fong2019understanding}
Ruth Fong, Mandela Patrick, and Andrea Vedaldi.
\newblock Understanding deep networks via extremal perturbations and smooth masks.
\newblock In \emph{ICCV}, pages 2950--2958, 2019.

\bibitem[Fong and Vedaldi(2017)]{fong2017interpretable}
Ruth~C Fong and Andrea Vedaldi.
\newblock Interpretable explanations of black boxes by meaningful perturbation.
\newblock In \emph{ICCV}, pages 3429--3437, 2017.

\bibitem[Fu et~al.(2020)Fu, Hu, Dong, Guo, Gao, and Li]{fu2020axiom}
Ruigang Fu, Qingyong Hu, Xiaohu Dong, Yulan Guo, Yinghui Gao, and Biao Li.
\newblock Axiom-based grad-cam: Towards accurate visualization and explanation of cnns.
\newblock In \emph{BMVC}, pages 1--13, 2020.

\bibitem[Hama et~al.(2023)Hama, Mase, and Owen]{hama2023deletion}
Naofumi Hama, Masayoshi Mase, and Art~B Owen.
\newblock Deletion and insertion tests in regression models.
\newblock \emph{Journal of Machine Learning Research}, 24\penalty0 (290):\penalty0 1--38, 2023.

\bibitem[He et~al.(2015)He, Zhang, Ren, and Sun]{he2015delving}
Kaiming He, Xiangyu Zhang, Shaoqing Ren, and Jian Sun.
\newblock Delving deep into rectifiers: Surpassing human-level performance on imagenet classification.
\newblock In \emph{ICCV}, pages 1026--1034, 2015.

\bibitem[He et~al.(2016)He, Zhang, Ren, and Sun]{he2016deep}
Kaiming He, Xiangyu Zhang, Shaoqing Ren, and Jian Sun.
\newblock Deep residual learning for image recognition.
\newblock In \emph{CVPR}, pages 770--778, 2016.

\bibitem[Jiang et~al.(2021)Jiang, Zhang, Hou, Cheng, and Wei]{jiang2021layercam}
Peng-Tao Jiang, Chang-Bin Zhang, Qibin Hou, Ming-Ming Cheng, and Yunchao Wei.
\newblock Layercam: Exploring hierarchical class activation maps for localization.
\newblock \emph{TIP}, 30:\penalty0 5875--5888, 2021.

\bibitem[Kingma(2014)]{kingma2014adam}
Diederik~P Kingma.
\newblock Adam: A method for stochastic optimization.
\newblock \emph{arXiv preprint arXiv:1412.6980}, 2014.

\bibitem[Klein et~al.(2024)Klein, L{\"u}th, Schlegel, Bungert, El-Assady, and J{\"a}ger]{klein2024navigating}
Lukas Klein, Carsten~T L{\"u}th, Udo Schlegel, Till~J Bungert, Mennatallah El-Assady, and Paul~F J{\"a}ger.
\newblock Navigating the maze of explainable ai: A systematic approach to evaluating methods and metrics.
\newblock \emph{arXiv preprint arXiv:2409.16756}, 2024.

\bibitem[Krause et~al.(2013)Krause, Stark, Deng, and Fei-Fei]{krause20133d}
Jonathan Krause, Michael Stark, Jia Deng, and Li Fei-Fei.
\newblock 3d object representations for fine-grained categorization.
\newblock In \emph{ICCVW}, pages 554--561, 2013.

\bibitem[Lundberg and Lee(2017{\natexlab{a}})]{lundberg2017unified}
Scott~M. Lundberg and Su-In Lee.
\newblock A unified approach to interpreting model predictions.
\newblock In \emph{NeurIPS}, page 4768–4777, 2017{\natexlab{a}}.

\bibitem[Lundberg and Lee(2017{\natexlab{b}})]{lundberg_unified_2017}
Scott~M Lundberg and Su-In Lee.
\newblock A {Unified} {Approach} to {Interpreting} {Model} {Predictions}.
\newblock In \emph{NeurIPS}, 2017{\natexlab{b}}.

\bibitem[Maji et~al.(2013)Maji, Rahtu, Kannala, Blaschko, and Vedaldi]{maji2013fine}
Subhransu Maji, Esa Rahtu, Juho Kannala, Matthew Blaschko, and Andrea Vedaldi.
\newblock Fine-grained visual classification of aircraft.
\newblock \emph{arXiv preprint arXiv:1306.5151}, 2013.

\bibitem[Mehrab et~al.(2024)Mehrab, Maruf, Daw, Manogaran, Neog, Khurana, Altintas, Bakis, Campolongo, Thompson, et~al.]{mehrab2024fish}
Kazi~Sajeed Mehrab, M Maruf, Arka Daw, Harish~Babu Manogaran, Abhilash Neog, Mridul Khurana, Bahadir Altintas, Yasin Bakis, Elizabeth~G Campolongo, Matthew~J Thompson, et~al.
\newblock Fish-vista: A multi-purpose dataset for understanding \& identification of traits from images.
\newblock \emph{arXiv preprint arXiv:2407.08027}, 2024.

\bibitem[Minh et~al.(2022)Minh, Wang, Li, and Nguyen]{minh2022explainable}
Dang Minh, H~Xiang Wang, Y~Fen Li, and Tan~N Nguyen.
\newblock Explainable artificial intelligence: a comprehensive review.
\newblock \emph{Artificial Intelligence Review}, pages 1--66, 2022.

\bibitem[Muhammad and Yeasin(2020)]{muhammad2020eigen}
Mohammed~Bany Muhammad and Mohammed Yeasin.
\newblock Eigen-cam: Class activation map using principal components.
\newblock In \emph{IJCNN}, pages 1--7. IEEE, 2020.

\bibitem[Oh et~al.(2021)Oh, Jung, Park, and Kim]{oh2021evet}
Youngrock Oh, Hyungsik Jung, Jeonghyung Park, and Min~Soo Kim.
\newblock Evet: enhancing visual explanations of deep neural networks using image transformations.
\newblock In \emph{WACV}, pages 3579--3587, 2021.

\bibitem[Oquab et~al.(2023)Oquab, Darcet, Moutakanni, Vo, Szafraniec, Khalidov, Fernandez, Haziza, Massa, El-Nouby, et~al.]{oquab2023dinov2}
Maxime Oquab, Timoth{\'e}e Darcet, Th{\'e}o Moutakanni, Huy Vo, Marc Szafraniec, Vasil Khalidov, Pierre Fernandez, Daniel Haziza, Francisco Massa, Alaaeldin El-Nouby, et~al.
\newblock Dinov2: Learning robust visual features without supervision.
\newblock \emph{arXiv preprint arXiv:2304.07193}, 2023.

\bibitem[Paul et~al.(2024)Paul, Chowdhury, Xiong, Chang, Carlyn, Stevens, Provost, Karpatne, Carstens, Rubenstein, et~al.]{paul2024simple}
Dipanjyoti Paul, Arpita Chowdhury, Xinqi Xiong, Feng-Ju Chang, David~Edward Carlyn, Samuel Stevens, Kaiya~L Provost, Anuj Karpatne, Bryan Carstens, Daniel Rubenstein, et~al.
\newblock A simple interpretable transformer for fine-grained image classification and analysis.
\newblock In \emph{ICLR}, 2024.

\bibitem[Petsiuk et~al.(2018)Petsiuk, Das, and Saenko]{petsiuk2018rise}
V Petsiuk, A Das, and K Saenko.
\newblock Rise: Randomized input sampling for explanation of black-box models.
\newblock In \emph{BMVC}, pages 1--13, 2018.

\bibitem[Piosenka(2023)]{piosenka2023birds}
Gerald Piosenka.
\newblock Birds 525 species - image classification.
\newblock 2023.

\bibitem[Radford et~al.(2021)Radford, Kim, Hallacy, Ramesh, Goh, Agarwal, Sastry, Askell, Mishkin, Clark, et~al.]{radford2021learning}
Alec Radford, Jong~Wook Kim, Chris Hallacy, Aditya Ramesh, Gabriel Goh, Sandhini Agarwal, Girish Sastry, Amanda Askell, Pamela Mishkin, Jack Clark, et~al.
\newblock Learning transferable visual models from natural language supervision.
\newblock In \emph{ICML}, pages 8748--8763. PMLR, 2021.

\bibitem[Ramaswamy et~al.(2020)]{ramaswamy2020ablation}
Harish~Guruprasad Ramaswamy et~al.
\newblock Ablation-cam: Visual explanations for deep convolutional network via gradient-free localization.
\newblock In \emph{WACV}, pages 983--991, 2020.

\bibitem[Rebuffi et~al.(2020)Rebuffi, Fong, Ji, and Vedaldi]{rebuffi2020there}
Sylvestre-Alvise Rebuffi, Ruth Fong, Xu Ji, and Andrea Vedaldi.
\newblock There and back again: Revisiting backpropagation saliency methods.
\newblock In \emph{CVPR}, pages 8839--8848, 2020.

\bibitem[Ribeiro et~al.(2016)Ribeiro, Singh, and Guestrin]{ribeiro2016should}
Marco~Tulio Ribeiro, Sameer Singh, and Carlos Guestrin.
\newblock " why should i trust you?" explaining the predictions of any classifier.
\newblock In \emph{KDD}, pages 1135--1144, 2016.

\bibitem[Ridnik et~al.(2021)Ridnik, Ben-Baruch, Noy, and Zelnik-Manor]{ridnik2021imagenet21k}
Tal Ridnik, Emanuel Ben-Baruch, Asaf Noy, and Lihi Zelnik-Manor.
\newblock Imagenet-21k pretraining for the masses, 2021.

\bibitem[Selvaraju et~al.(2017)Selvaraju, Cogswell, Das, Vedantam, Parikh, and Batra]{selvaraju2017grad}
Ramprasaath~R Selvaraju, Michael Cogswell, Abhishek Das, Ramakrishna Vedantam, Devi Parikh, and Dhruv Batra.
\newblock Grad-cam: Visual explanations from deep networks via gradient-based localization.
\newblock In \emph{ICCV}, pages 618--626, 2017.

\bibitem[Smilkov et~al.(2017)Smilkov, Thorat, Kim, Vi{\'e}gas, and Wattenberg]{smilkov2017smoothgrad}
Daniel Smilkov, Nikhil Thorat, Been Kim, Fernanda Vi{\'e}gas, and Martin Wattenberg.
\newblock Smoothgrad: removing noise by adding noise.
\newblock \emph{arXiv preprint arXiv:1706.03825}, 2017.

\bibitem[Szegedy et~al.(2015)Szegedy, Liu, Jia, Sermanet, Reed, Anguelov, Erhan, Vanhoucke, and Rabinovich]{szegedy2015going}
Christian Szegedy, Wei Liu, Yangqing Jia, Pierre Sermanet, Scott Reed, Dragomir Anguelov, Dumitru Erhan, Vincent Vanhoucke, and Andrew Rabinovich.
\newblock Going deeper with convolutions.
\newblock In \emph{CVPR}, pages 1--9, 2015.

\bibitem[Wah et~al.(2011)Wah, Branson, Welinder, Perona, and Belongie]{wah2011caltech}
Catherine Wah, Steve Branson, Peter Welinder, Pietro Perona, and Serge Belongie.
\newblock The caltech-ucsd birds-200-2011 dataset.
\newblock 2011.

\bibitem[Wang et~al.(2020)Wang, Wang, Du, Yang, Zhang, Ding, Mardziel, and Hu]{wang2020score}
Haofan Wang, Zifan Wang, Mengnan Du, Fan Yang, Zijian Zhang, Sirui Ding, Piotr Mardziel, and Xia Hu.
\newblock Score-cam: Score-weighted visual explanations for convolutional neural networks.
\newblock In \emph{CVPRW}, pages 24--25, 2020.

\bibitem[Wei et~al.(2021)Wei, Song, Mac~Aodha, Wu, Peng, Tang, Yang, and Belongie]{wei2021fine}
Xiu-Shen Wei, Yi-Zhe Song, Oisin Mac~Aodha, Jianxin Wu, Yuxin Peng, Jinhui Tang, Jian Yang, and Serge Belongie.
\newblock Fine-grained image analysis with deep learning: A survey.
\newblock \emph{TPAMI}, 44\penalty0 (12):\penalty0 8927--8948, 2021.

\bibitem[Xue et~al.(2022)Xue, Huang, Zhang, Cheng, Song, Wu, and Song]{xue2022protopformer}
Mengqi Xue, Qihan Huang, Haofei Zhang, Lechao Cheng, Jie Song, Minghui Wu, and Mingli Song.
\newblock Protopformer: Concentrating on prototypical parts in vision transformers for interpretable image recognition.
\newblock \emph{arXiv preprint arXiv:2208.10431}, 2022.

\bibitem[Yeh et~al.(2019)Yeh, Hsieh, Suggala, Inouye, and Ravikumar]{yeh2019fidelity}
Chih-Kuan Yeh, Cheng-Yu Hsieh, Arun~Sai Suggala, David~I Inouye, and Pradeep Ravikumar.
\newblock On the (in) fidelity and sensitivity of explanations.
\newblock In \emph{NeurIPS}, pages 10967--10978, 2019.

\bibitem[Zhao et~al.(2017)Zhao, Feng, Wu, and Yan]{zhao2017survey}
Bo Zhao, Jiashi Feng, Xiao Wu, and Shuicheng Yan.
\newblock A survey on deep learning-based fine-grained object classification and semantic segmentation.
\newblock \emph{International Journal of Automation and Computing}, 14\penalty0 (2):\penalty0 119--135, 2017.

\bibitem[Zheng et~al.(2017)Zheng, Fu, Mei, and Luo]{zheng2017learning}
Heliang Zheng, Jianlong Fu, Tao Mei, and Jiebo Luo.
\newblock Learning multi-attention convolutional neural network for fine-grained image recognition.
\newblock In \emph{ICCV}, pages 5209--5217, 2017.

\bibitem[Zheng et~al.(2019)Zheng, Fu, Zha, and Luo]{zheng2019looking}
Heliang Zheng, Jianlong Fu, Zheng-Jun Zha, and Jiebo Luo.
\newblock Looking for the devil in the details: Learning trilinear attention sampling network for fine-grained image recognition.
\newblock In \emph{CVPR}, pages 5012--5021, 2019.

\bibitem[Zhou et~al.(2016)Zhou, Khosla, Lapedriza, Oliva, and Torralba]{zhou2016learning}
Bolei Zhou, Aditya Khosla, Agata Lapedriza, Aude Oliva, and Antonio Torralba.
\newblock Learning deep features for discriminative localization.
\newblock In \emph{CVPR}, pages 2921--2929, 2016.

\end{thebibliography}
}

\clearpage
\setcounter{page}{1}
\maketitlesupplementary
\appendix

% \section*{Appendix}

\noindent
The supplementary material is organized into the following sections.
\cref{sec:app-method-detail} provides more details of the method implementation and experiment settings. \cref{sec:app-experiment} discusses experimental results on more datasets and model architectures, and \cref{sec:app-vis} presents more visualizations of the proposed \Ours.

\section{More Implementation Details}\label{sec:app-method-detail}
\subsection{Sorted Weight Similarity Distribution}
We show the distribution of sorted weight similarity of three datasets in~\cref{fig:teaser}. 
Here we provide the implementation details for reproducing the curves. 
First, we train a linear classifier for each dataset on top of the pre-trained CLIP visual encoder~\cite{radford2021learning}. 
The visual encoder is frozen during the classifier training.
Next, we calculate the similarity matrix $\mS$ between the weights of the linear classifier with each element $S_{pq}$ defined by:
\begin{equation}
    S_{pq}=\frac{\vw^p\cdot \vw^q}{\Vert \vw^p\Vert_2\Vert \vw^q\Vert_2},
\end{equation}
where $\vw^p$ and $\vw^q$ represent the linear classifier weights for class $p$ and class $q$, respectively. 
The diagonal elements are subtracted by 1 to eliminate self-similarity. 
The similarity matrix is then sorted in descending order for each class:
\begin{equation}
    \mS^{\text{sorted}}=\text{sort\_rows}(\mS),
\end{equation}
such that the first element of each row has the largest similarity to the query class. 
Last, we compute the class-wise average of the sorted similarity values to generate the distribution curve. The shaded regions in the figure stand for standard deviation. 
Therefore, the leftmost point of each curve reflects the average similarity between one class and its most similar counterpart. 
Although after model training, the average similarity is low, for each class, there still exist certain other classes with high similarities. 

\subsection{Activation Faithfulness Examination}
Based on the extension to multi-modal zero-shot models, the proposed \Ours can be applied to verify if the prediction of a linear classifier faithfully aligns with real class attributes, as illustrated in~\cref{sec:analysis}. 
Here we provide more implementation details of the process. 
The CUB-200 dataset~\cite{wah2011caltech} provides continuous attribute labels for each class. 
Given one target class, we conduct subtraction between the attribute labels of the target class and those of the most similar class. 
The top 3 attributes with the largest value difference are selected as discriminative attributes, and are to be highlighted in the image. 

Next, we generate two saliency maps for one given image of the specified class. 
The first saliency map is obtained based on~\cref{eq:finercam} to maximize the difference between the target class and similar class prediction logits. 
It reflects the distinctions recognized by the classifier model. 
The second saliency map is obtained by setting text attribute labels and the general category ``bird'' as comparing pairs in the zero-shot classification setting. 
It shows the ``ground truth'' difference between the two classes. 
Subsequently, we can compare if the classifier-based saliency map activates similar regions as the attribute-based one. 
An aligned saliency map pair indicates that the classifier is looking at correct hints to distinguish the image. 
Oppositely, if the saliency maps misalign, either the classifier is not working as expected, or there are certain traits not labeled by the dataset. 

\subsection{Dataset Information}

We utilized five publicly available datasets to evaluate our method. Below, we summarize the key characteristics of each dataset, including the number of categories, sample distributions, and additional details provided by the respective dataset sources.

\begin{table}[t]
    \caption{Classification accuracy (\%) of linear probing on DINOv2 and CLIP backbones on five datasets.}
    \label{tab:model-accuracy}
    \centering
    \small
    \vskip -4pt
    \resizebox{\columnwidth}{!}{
    \begin{tabular}{c|ccccc}
    \toprule
        Model & Birds-525 & CUB-200 & Cars & Aircraft & FishVista \\
    \midrule
        CLIP &
        95.3 & 58.4 & 64.9 & 53.9 & 64.6 \\
        DINOv2 &
        97.5 & 66.4 & 78.7 & 83.9 & 79.6 \\
    \bottomrule
    \end{tabular}
    }
    \vskip -4pt
\end{table}

\begin{table*}[t]
    \caption{The quantitative evaluation results on the proposed Finer-CAM and baseline CAM methods on FishVista and Aircraft. The abbreviations stand for deletion, relative drop, and localization, respectively. }
    \label{tab:app-add-metric}
    \centering
    \small
    \resizebox{0.7\linewidth}{!}{
    \begin{tabular}{l|ccc|cccc}
    \toprule
        \multirow{2}{*}{Method} & \multicolumn{3}{c|}{FishVista} & \multicolumn{4}{c}{Aircraft} \\
        & Del. $\downarrow$& RD.$_{\text{@}0.05}$ $\uparrow$ & RD.$_{\text{@}0.1}$ $\uparrow$ & Del. $\downarrow$ & RD.$_{\text{@}0.05}$ $\uparrow$ & RD.$_{\text{@}0.1}$ $\uparrow$ & Loc. $\uparrow$ \\
    \midrule
        Grad-CAM~\cite{selvaraju2017grad} &
        \textbf{0.037} & 0.177 & 0.205 &
        \textbf{0.039} & 0.097 & 0.112 &\ 0.608 \\
        + Finer &
        0.039 & \textbf{0.193} & \textbf{0.217} &
        \textbf{0.039} & \textbf{0.113} & \textbf{0.127} &\textbf{0.614} \\
    \midrule
        Layer-CAM~\cite{jiang2021layercam} &
        \textbf{0.049} & 0.163 & 0.181 &
        \textbf{0.037} & 0.101 & 0.113 & \ 0.662\\
        + Finer &
        \textbf{0.049} & \textbf{0.196} & \textbf{0.210} &
        0.039 & \textbf{0.113} & \textbf{0.124} &\textbf{0.664} \\
    \midrule
        Score-CAM~\cite{wang2020score} &
        \textbf{0.051} & 0.158 & 0.188 &
        \textbf{0.050} & 0.074 & 0.086 &\ 0.595 \\
        + Finer &
        0.052 & \textbf{0.174} & \textbf{0.203} &
        \textbf{0.050} & \textbf{0.085} & \textbf{0.094} &\textbf{0.602} \\
    \bottomrule
    \end{tabular}
    }
\end{table*}

\begin{table*}[t]
    \caption{The quantitative evaluation results on the proposed Finer-CAM and baseline CAM methods \textbf{with DINOv2 as the backbone}. The abbreviations stand for deletion, relative drop, and localization, respectively. }
    \label{tab:app-dino-metric}
    \centering
    \small
    \resizebox{\linewidth}{!}{
    \begin{tabular}{l|ccc|cccc|cccc}
    \toprule
        \multirow{2}{*}{Method} & \multicolumn{3}{c|}{Birds525} & \multicolumn{4}{c|}{CUB} & \multicolumn{4}{c}{Cars} \\
        & Del. $\downarrow$& RD.$_{\text{@}0.05}$ $\uparrow$ & RD.$_{\text{@}0.1}$ $\uparrow$ & Del. $\downarrow$ & RD.$_{\text{@}0.05}$ $\uparrow$ & RD.$_{\text{@}0.1}$ $\uparrow$ & Loc. $\uparrow$ & Del. $\downarrow$ & RD.$_{\text{@}0.05}$ $\uparrow$ & RD.$_{\text{@}0.1}$ $\uparrow$ & Loc. $\uparrow$ \\
    \midrule
        Grad-CAM~\cite{selvaraju2017grad} &
        0.252 & 0.041 & 0.069 &
        0.171 & 0.124 & 0.157 & 0.500&
        \textbf{0.088} & 0.222 & 0.280 & \ 0.619\\
        + Finer &
        \textbf{0.250} & \textbf{0.049} & \textbf{0.080} &
        \textbf{0.165} & \textbf{0.151} & \textbf{0.185} &\textbf{0.530} &
        0.091 & \textbf{0.243} & \textbf{0.306} & \textbf{0.632}\\
    \midrule
        Layer-CAM~\cite{jiang2021layercam} &
        \textbf{0.254} & 0.047 & 0.075 &
        \textbf{0.143} & 0.174 & 0.210 & 0.682 &
        \textbf{0.105} & 0.210 & 0.270 & \ 0.618 \\
        + Finer &
        0.258 & \textbf{0.055} & \textbf{0.079} &
        0.148 & \textbf{0.192} & \textbf{0.230} &\textbf{0.729} &
        0.108 & \textbf{0.236} & \textbf{0.294} & \textbf{0.647} \\
    \midrule
        Score-CAM~\cite{wang2020score} &
        \textbf{0.282} & \textbf{0.042} & 0.062 &
        \textbf{0.174} & 0.125 & 0.157 & 0.630 &
        \textbf{0.152} & 0.127 & 0.173 & \ 0.579\\
        + Finer &
        0.284 & 0.036 & \textbf{0.064} &
        0.176 & \textbf{0.137} & \textbf{0.168} &\textbf{0.640} &
        \textbf{0.152} & \textbf{0.141} & \textbf{0.191} &\textbf{0.586}  \\
    \bottomrule
    \end{tabular}
    }
\end{table*}

\begin{table*}[t!]
    \caption{The quantitative evaluation results on the proposed Finer-CAM and baseline CAM methods on FishVista and Aircraft \textbf{with DINOv2 as the backbone}. The abbreviations stand for deletion, relative drop, and localization, respectively. }
    \label{tab:app-dino-metric-fish}
    \centering
    \small
    \resizebox{0.7\linewidth}{!}{
    \begin{tabular}{l|ccc|cccc}
    \toprule
        \multirow{2}{*}{Method} & \multicolumn{3}{c|}{FishVista} & \multicolumn{4}{c}{Aircraft} \\
        & Del. $\downarrow$& RD.$_{\text{@}0.05}$ $\uparrow$ & RD.$_{\text{@}0.1}$ $\uparrow$ & Del. $\downarrow$ & RD.$_{\text{@}0.05}$ $\uparrow$ & RD.$_{\text{@}0.1}$ $\uparrow$ & Loc. $\uparrow$ \\
    \midrule
        Grad-CAM~\cite{selvaraju2017grad} &
         \textbf{0.132} & 0.206 & 0.270 &
         \textbf{0.178} & 0.242 & 0.309 & \ 0.561\\
        + Finer &
         0.135 & \textbf{0.224} & \textbf{0.290} &
         \textbf{0.178} & \textbf{0.270} & \textbf{0.339} & \textbf{0.586}\\
    \midrule
        Layer-CAM~\cite{jiang2021layercam} &
        \textbf{0.129} & 0.215 & 0.278 &
        \textbf{0.168} & 0.286 & 0.367 & \ 0.729\\
        + Finer &
        0.134 & \textbf{0.220} & \textbf{0.288} &
        0.170 & \textbf{0.312} & \textbf{0.383} & \textbf{0.749}\\
    \midrule
        Score-CAM~\cite{wang2020score} &
        \textbf{0.154} & 0.159 & 0.210 &
        \textbf{0.198} & 0.182 & 0.257 & \ 0.611\\
        + Finer &
        0.159 & \textbf{0.173} &  \textbf{0.229} &
        0.203 & \textbf{0.194} & \textbf{0.264} &\textbf{0.653} \\
    \bottomrule
    \end{tabular}
    }
\end{table*}

\begin{itemize}
    \item \textbf{Birds-525}~\cite{piosenka2023birds}: 
    This dataset comprises 525 bird species with 84,635 training images and 2,625 validation images. It provides a diverse collection suitable for fine-grained classification tasks.

    \item \textbf{CUB-200}~\cite{wah2011caltech}: 
    This dataset is a benchmark for fine-grained categorization with 11,788 images across 200 bird species. The dataset includes 5,994 training images and 5,794 testing images, with detailed annotations such as subcategory labels, part locations, and bounding boxes.

    \item \textbf{Cars}~\cite{krause20133d}: 
    This dataset contains 16,185 images of 196 car models, split into 8,144 training images and 8,041 testing images. Categories include make, model, and year, making it ideal for subtle visual recognition tasks.

    \item \textbf{Aircraft}~\cite{maji2013fine}: 
    This dataset comprises 10,200 aircraft images annotated across 70 family-level categories. The dataset is divided into training, validation, and test subsets of 3,334 images each, with hierarchical annotations for classification.

    \item \textbf{FishVista}~\cite{mehrab2024fish}: 
    This dataset is a large collection of 60,000 fish images spanning 1,900 species, designed for species classification and trait identification. We use a subset of 414 species, with 35,328 training images, 4,996 validation images, and 7,556 test images. It includes fine-grained annotations and pixel-level segmentations for 2,427 images.
\end{itemize}

\section{More Experimental Results}\label{sec:app-experiment}
\subsection{Model Accuracy}

We present the classification accuracy of linear probing on two backbones, CLIP~\cite{radford2021learning} and DINOv2~\cite{oquab2023dinov2}, on five datasets~\cref{tab:model-accuracy} summarizes the results.
Generally, DINOv2 provides visual embeddings with better quality and achieves higher classification accuracy. 
We use OpenCLIP ViT-B-16 (pre-trained on LAION-400M) in all the experiments.

\subsection{Results on FishVista and Aircraft}
In addition to~\cref{tab:metric}, we also conduct the quantitative evaluation on the FishVista~\cite{mehrab2024fish} and Aircraft~\cite{krause20133d} datasets in~\cref{tab:app-add-metric}. 
\Ours yields similar performance on the deletion AUC as baselines while performing much better in terms of relative drop and localization metrics. 
The performance superiority further supports the effectiveness of the proposed \Ours method.

\subsection{Results on DINOv2}
We adopt the pre-trained CLIP model~\cite{radford2021learning} as the backbone in the previous experiments. 
Here, we further employ DINOv2~\cite{oquab2023dinov2} to extract visual embeddings for generating saliency maps. 
We report the results on the five adopted datasets in~\cref{tab:app-dino-metric} and~\cref{tab:app-dino-metric-fish}. 
Similarly, the proposed \Ours achieves higher relative drop and localization performance compared with baselines. 
It indicates that Fine-CAM can be applied to a variety of architectures and provide effective interpretation.
It can be observed in~\cref{tab:model-accuracy} that the linear classifier trained on top of DINOv2 achieves higher accuracy than that on CLIP. 
As a result, it requires deleting more pixels to decrease the confidence of the target class, leading to larger deletion AUC values and a smaller relative drop in some cases compared with CLIP. 

\begin{table}[t]
    \caption{The comparison of different aggregation strategies. Del. and RD.$_{\text{@}0.05}$ represent deletion AUC and relative drop when masking out the top 5\% activated pixels, respectively.}
    \label{tab:aggregation}
    \centering
    \small
    % \resizebox{\columnwidth}{!}{
    \begin{tabular}{l|cc|c}
    \toprule
       \multirow{2}{*}{Aggregation} & \multicolumn{2}{c|}{Before \texttt{ReLU}} & \multicolumn{1}{c}{After \texttt{ReLU}} \\ \cline{2-4}
       & Max & Avg & Avg \\
    \midrule
        Del. $\downarrow$ & 0.081 & \textbf{0.080} & 0.081  \\
        RD.$_{\text{@}0.05}$ $\uparrow$ & 0.184 & \textbf{0.192} & 0.191 \\
    \bottomrule
    \end{tabular}
    % }
    \vskip -4pt
\end{table}

\subsection{Aggregation strategy.}
There are multiple potential strategies to aggregate the activations from different comparison references (cf.~\cref{eq:multiple-agg}). 
\cref{tab:aggregation} summarizes the comparison of three aggregation ways.
Generally, averaging the activation weights from different references before the \texttt{ReLU} operation yields the best performance. 

\section{More Visualizations}\label{sec:app-vis}

\subsection{Failure Cases}
\begin{figure}[t]
\centering
    \includegraphics[width=\columnwidth]{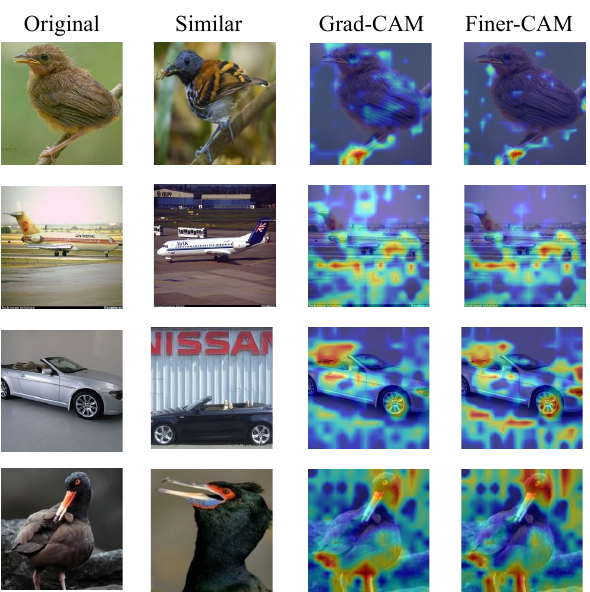}
    \caption{\textbf{Visualization of some failure cases} where \Ours cannot produce better saliency maps than the Grad-CAM baseline. }
    \label{fig:failure}
    \vskip -6pt
\end{figure}

We include some failure cases in~\cref{fig:failure}.
In these examples, the baseline Grad-CAM highlights large portions of the images that are not the target objects. 
Through further analysis, it often happens when the classifier fails to provide a correct prediction. Under these circumstances, \Ours also cannot interpret the decision effectively.
Finer-CAM may degenerate to baseline methods when the logit similarity does not reflect visual similarity, \ie, the target class is significantly different from others.

\subsection{Multi-modal Interaction}
\begin{figure*}[t]
\centering
    \includegraphics[width=\textwidth]{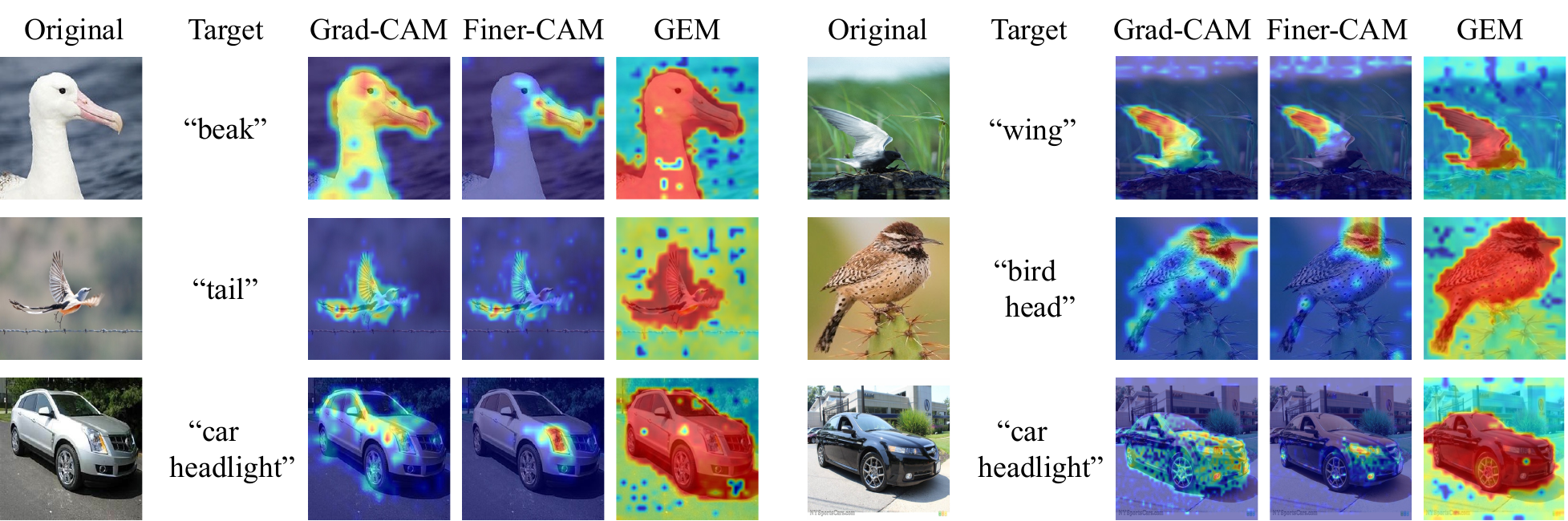}
    \caption{\textbf{Visualization of multi-modal localization of fine-grained traits or object parts}. For each original image, we aim to locate the target concept. By emphasizing the difference between the target concept and the original concept (``bird'' or ``car''), \Ours accurately localizes the target image regions. }
    \label{fig:multimodal}
    \vskip -6pt
\end{figure*}

We demonstrate that in addition to interpreting classifiers, the proposed \Ours can also be applied to multi-modal scenarios to localize concepts in the images.
We provide more examples in~\cref{fig:multimodal}.
Using Grad-CAM with the target concept alone often leads to inaccurate or wrong activations. 
In comparison, with a base concept (\textit{e.g.}, ``bird'' or ``car'') as reference, emphasizing their difference produces substantially more accurate localization of fine-grained traits or object parts.
We also compare the localization capability with a recent method GEM~\cite{bousselham2024grounding}.
GEM is capable of grounding the target object in the images.
However, when asked to localize fine-grained traits or object parts, it still yields activations over the entire object. 
\Ours, comparatively, is a better tool to highlight details.

\subsection{Qualitative Comparison}
\begin{figure*}[t]
\centering
    \includegraphics[width=\textwidth]{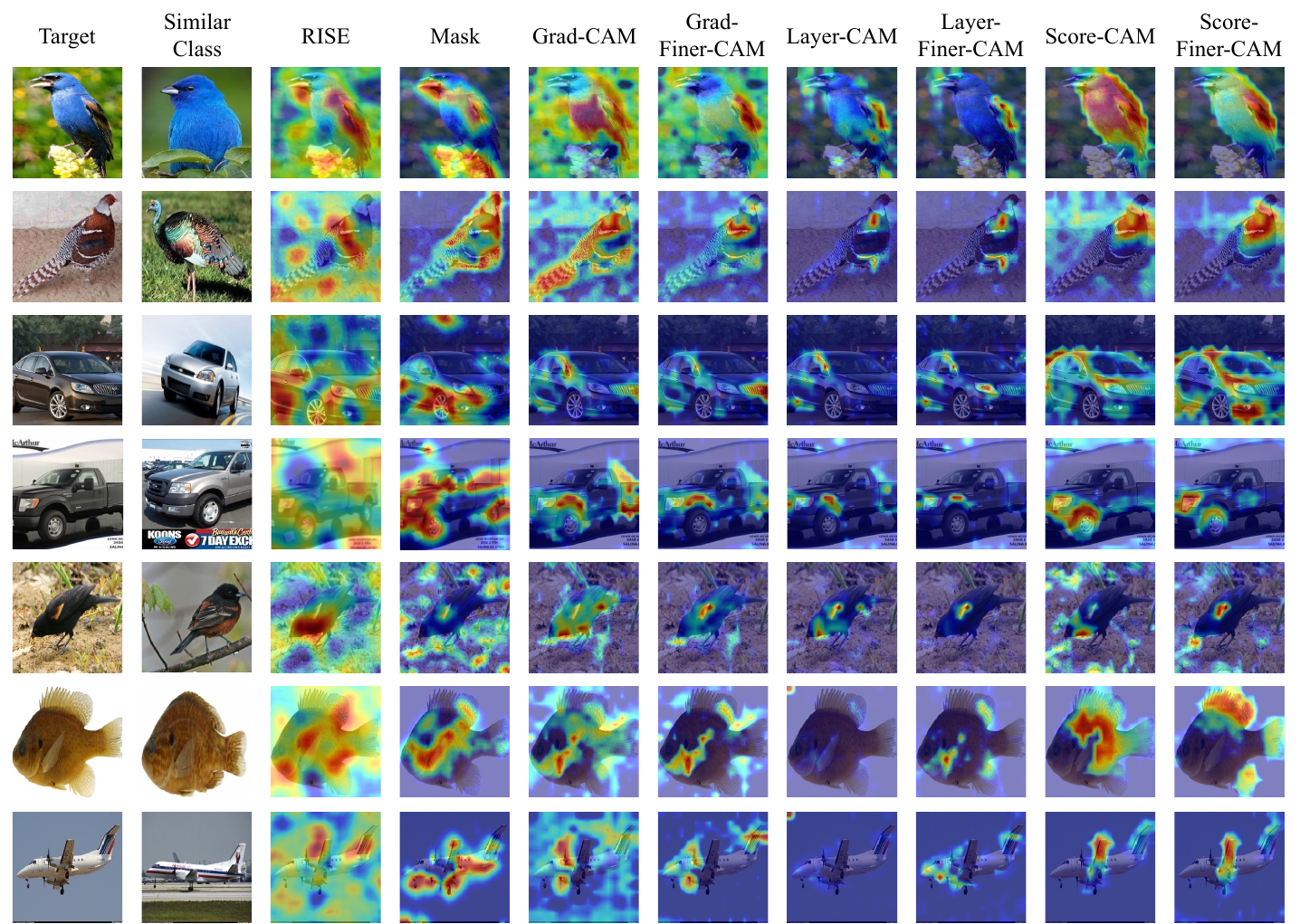}
    \caption{\textbf{Qualitative comparison of the saliency maps generated by baseline CAM methods} (Grad-CAM~\cite{selvaraju2017grad}, Layer-CAM~\cite{jiang2021layercam}, and Score-CAM~\cite{wang2020score}), the proposed \Ours applied on these three baselines, and other XAI methods (RISE~\cite{petsiuk2018rise} and Mask~\cite{fong2017interpretable}). }
    \label{fig:vis-more}
\end{figure*}

We visualize more examples in~\cref{fig:vis-more} on different datasets. The comparison also includes two XAI methods RISE~\cite{petsiuk2018rise} and Mask~\cite{fong2017interpretable}. 
The results are obtained with DINOv2~\cite{oquab2023dinov2} as the backbone. 
Comparatively, the proposed \Ours activates the most discriminative image regions that can tell the difference between the target class and similar classes, and also suppresses the noise in the background.

\subsection{Extrapolation}
\begin{figure*}[t]
\centering
    \includegraphics[width=0.7\textwidth]{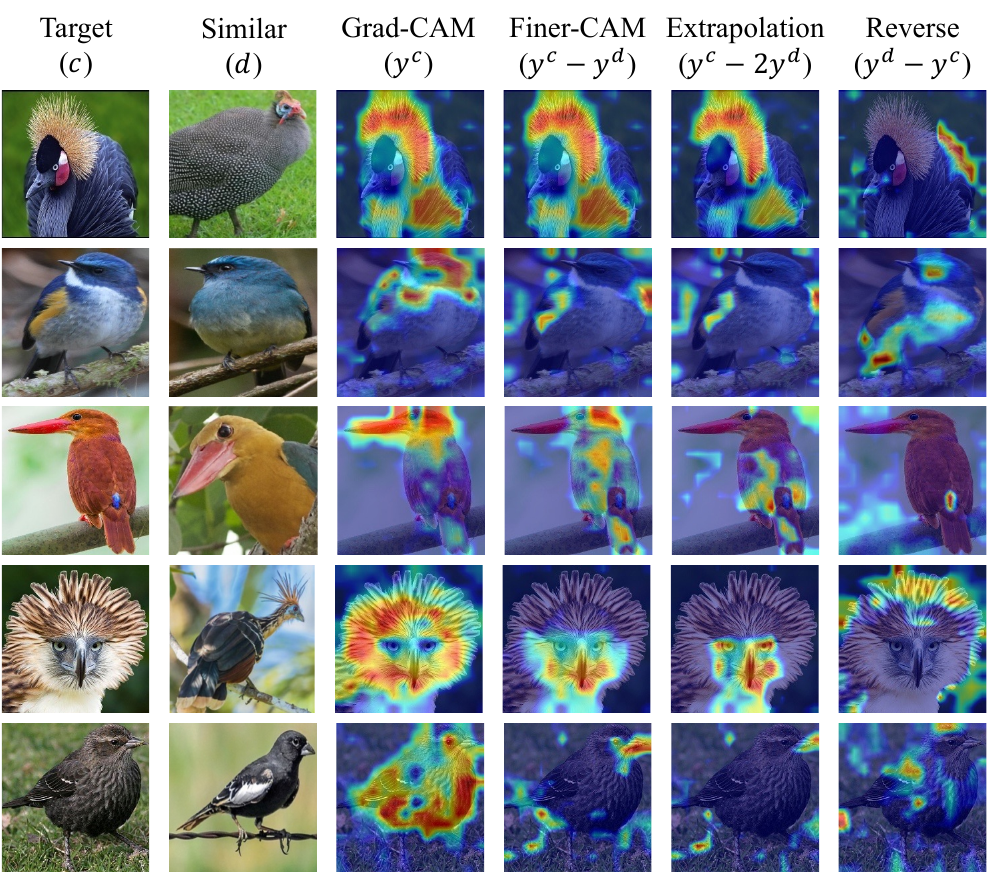}
    \caption{\textbf{Visualization of the extrapolation and reverse comparing cases with Grad-CAM as the baseline}. The first two columns show the target image from class $c$ and an image from the similar class $d$. The next two rows show the saliency maps generated by Grad-CAM and \Ours. \Ours calculates the gradients of the difference between two prediction logits to acquire the activation weights. \textbf{Extrapolation} further emphasizes the difference, while \textbf{Reverse} tries to look for the traits of the similar class in the target image. }
    \label{fig:reverse}
    \vskip -6pt
\end{figure*}

The proposed \Ours highlights those discriminative image regions that maximize the prediction difference $y^c-\gamma y^d$ between the target class $c$ and the similar class $d$. 
We have tested different $\gamma$ settings from 0.0 to 1.0 in the main text.
We also visualize the extrapolation case when $\gamma=2.0$ in~\cref{fig:reverse}.
Generally, the activations are more tensely highlighting subtle details. 

\subsection{Reverse Comparing}
The reverse comparing aims to look for features predictive of the similar class from the target image, which maximize $y^d-y^c$. 
The visualization examples are shown in the last column of~\cref{fig:reverse}.
The generated saliency maps can locate some traits that are predictive of the similar class, instead of the traits highlighted by \Ours.


\end{document}